\documentclass[lettersize,journal]{IEEEtran}
\usepackage{amsmath,amsfonts}
\usepackage{array}
\usepackage{textcomp}
\usepackage{stfloats}
\usepackage{url}
\usepackage{verbatim}
\usepackage{graphicx}
\usepackage{cite}
\usepackage[linesnumbered,ruled,vlined]{algorithm2e}
\usepackage{amsmath}
\usepackage{subcaption}
\usepackage{booktabs}
\usepackage{pifont}
\usepackage{multirow}
\usepackage{tabularray}
\usepackage{algpseudocode}  
\DeclareMathOperator*{\argmin}{arg\,min}

\begin{document}

\title{FedPAW: Federated Learning with Personalized Aggregation Weights for Urban Vehicle Speed Prediction}

\author{Yuepeng He, Pengzhan Zhou,~\IEEEmembership{Member,~IEEE}, Yijun Zhai, Fang Qu, Zhida Qin,~\IEEEmembership{Member,~IEEE}, Mingyan Li,~\IEEEmembership{Member,~IEEE}, Songtao Guo,~\IEEEmembership{Senior Member,~IEEE}
\thanks{This paper was produced by the XX. They are in XX.}
\thanks{Manuscript received XX; revised XX.}}

\markboth{IEEE TRANSACTIONS ON CLOUD COMPUTING, XX}%
{Shell \MakeLowercase{\textit{et al.}}: A Sample Article Using IEEEtran.cls for IEEE Journals}


\maketitle

\begin{abstract}

Vehicle speed prediction is crucial for intelligent transportation systems, promoting more reliable autonomous driving by accurately predicting future vehicle conditions. Due to variations in drivers' driving styles and vehicle types, speed predictions for different target vehicles may significantly differ. Existing methods may not realize personalized vehicle speed prediction while protecting drivers' data privacy. We propose a Federated learning framework with Personalized Aggregation Weights (FedPAW) to overcome these challenges. This method captures client-specific information by measuring the weighted mean squared error between the parameters of local models and global models. The server sends tailored aggregated models to clients instead of a single global model, without incurring additional computational and communication overhead for clients. To evaluate the effectiveness of FedPAW, we collected driving data in urban scenarios using the autonomous driving simulator CARLA, employing an LSTM-based Seq2Seq model with a multi-head attention mechanism to predict the future speed of target vehicles. The results demonstrate that our proposed FedPAW ranks lowest in prediction error within the time horizon of 10 seconds, with a 0.8\% reduction in test MAE, compared to eleven representative benchmark baselines. The source code of FedPAW and dataset CarlaVSP are open-accessed at: https://github.com/heyuepeng/PFLlibVSP and https://pan.baidu.com/s/1qs8fxUvSPERV3C9i6pfUIw?pwd=tl3e.
\end{abstract}


\section{Introduction}

\IEEEPARstart{V}{ehicle} speed prediction plays a significant role in modern intelligent transportation systems (ITS), serving as a critical technology to enhance road safety, traffic efficiency, and vehicle energy efficiency \cite{jiang2016vehicle}. It is widely applied in areas such as path planning, energy management of hybrid electric vehicles (HEVs) \cite{sun2014velocity, han2019short}, and ecological adaptive cruise control (EACC) \cite{moser2015short, jia2020lstm, chada2023deep}. However, accurately predicting the speed of an individual vehicle presents a considerable challenge, as vehicle speed can be influenced by traffic conditions, vehicle types, road conditions, and driver behaviors \cite{jiang2016vehicle}. Compared to traditional prediction methods \cite{bichi2010stochastic, zhang2011predictive, kumagai2006prediction, wang2014modeling}, Recurrent Neural Networks (RNNs) leveraging vehicle-to-vehicle (V2V) and vehicle-to-infrastructure (V2I) communication have been proven more effective for vehicle speed prediction \cite{han2019short, jia2020lstm, wegener2021longitudinal, chada2023deep}. Nevertheless, former researches have not accounted for the significant differences in drivers' driving styles and vehicle types, which greatly impact speed prediction for the target vehicle, hence failing to achieve personalized predictions. Collaborative training on driving data collected from multiple vehicles is crucial for enhancing prediction accuracy, as a single vehicle would not possess sufficient data to train a reliable model. However, traditional distributed machine learning techniques necessitate centralizing the private driving data of all vehicles on a central server (e.g., cloud servers), posing potential risks for private data leakage.

\begin{figure}[htpb]
    \centering
    \includegraphics[width=\linewidth]{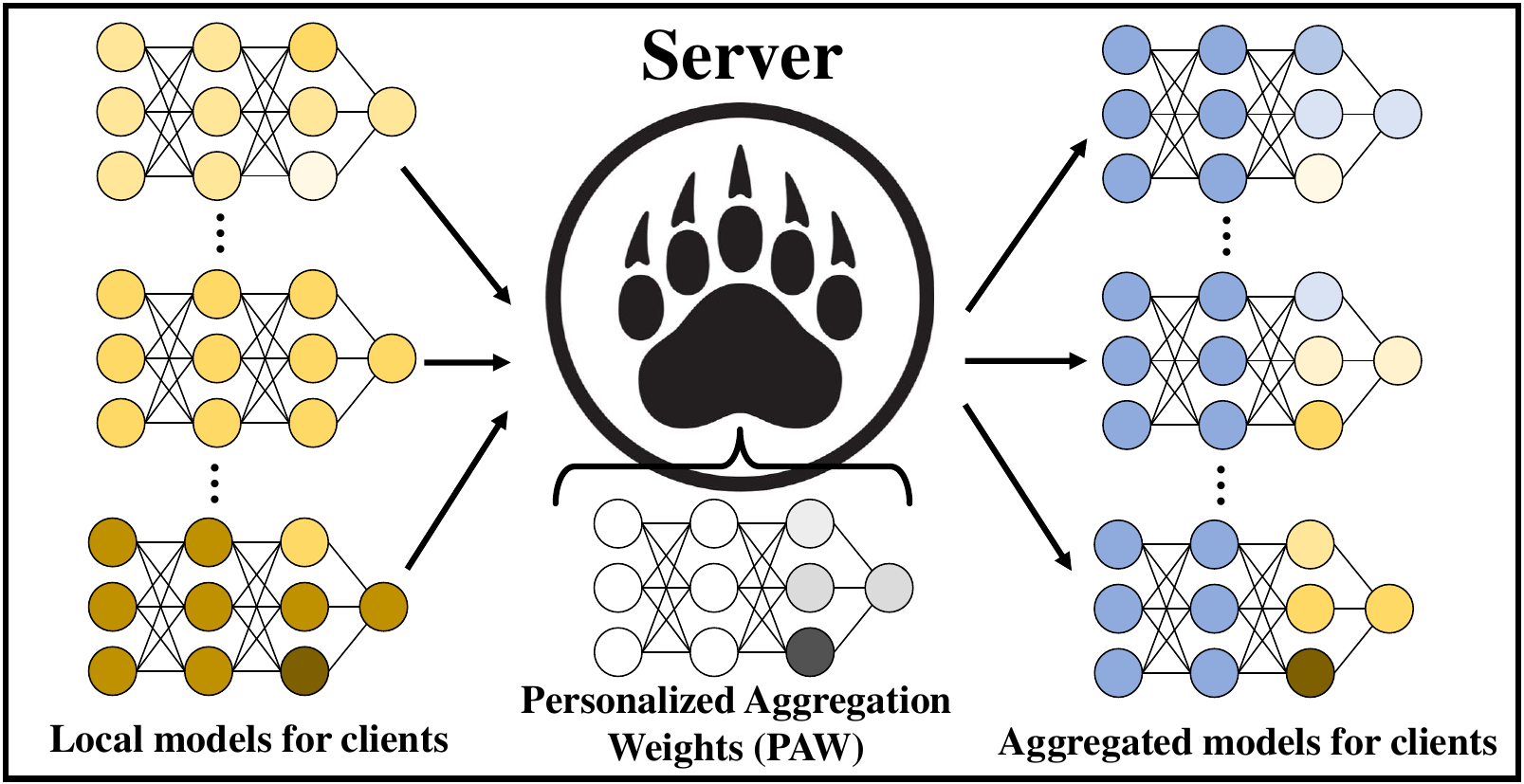}
    \caption{Illustration of FedPAW on the server.}
    \label{fig:PAW}
\end{figure}

Federated Learning (FL) emerges as a collaborative, distributed machine learning paradigm coordinated by a central server and solved jointly through a network of multiple participating devices (clients). Each client possesses its private dataset, not uploaded to the server. Instead, locally trained models are uploaded to aggregate into a global model on the server, effectively reducing privacy risks \cite{mcmahan2017communication}. However, traditional FL approaches like FedAvg \cite{mcmahan2017communication} and FedProx \cite{li2020federated}, when applied to vehicle speed prediction, fail to deliver personalized speed predictions with a single global model.

Personalized Federated Learning (PFL) methods address the statistical heterogeneity in FL, offering personalized solutions \cite{tan2022towards}. Unlike traditional FL, which seeks a single high-quality global model, PFL methods prioritize training local models for each client \cite{zhang2023fedala}. However, existing PFL methods almost invariably require additional steps on the client side to achieve personalization, often increasing the computational and communication overhead for clients. Given the real-time requirements of driving scenarios and the limited computational and communication resources of vehicles, implementing personalization steps on a powerful server appears to be more appropriate.

We propose a Federated learning framework with Personalized Aggregation Weights (FedPAW) to achieve personalized vehicle speed prediction. As shown in Figure\ref{fig:PAW}, FedPAW distinguishes itself from other PFL methods by implementing a Personalized Aggregation step on the server to capture client-specific information: aggregated models are obtained through element-wise aggregation of client local models and the global model. Aggregation weights are derived by calculating the weighted mean squared error between the parameters of client local models and the global model. The server can then distribute tailored aggregated models (i.e., personalized models) to each client, ensuring personalized prediction of vehicle speed without adding extra computational and communication overhead. Note that, a \emph{PAW} is depicted in Figure \ref{fig:PAW} to describe our method, since the lengths and functions of toes (similar to aggregation weights) of an animal's paw are different, and they together form a complete paw just as forming a complete personalized model. 

To evaluate the effectiveness of the FedPAW framework, we collected driving data from various drivers and vehicles in urban scenarios using the CARLA simulator \cite{dosovitskiy2017carla} and conducted extensive experiments on this dataset. Experimental results show that FedPAW outperforms eleven benchmark baselines, including local training, centralized training, FedAvg, FedProx, and seven state-of-the-art (SOTA) PFL methods \cite{fallah2020personalized, collins2021exploiting, t2020personalized, li2021ditto, deng2020adaptive, zhang2020personalized, zhang2023fedala}. In summary, our contributions are three-fold:
\begin{itemize}

\item We have collected and open-sourced a simulated driving dataset, CarlaVSP, through the CARLA simulator. This dataset distinguishes between different drivers and types of vehicles, providing benefits for subsequent personalized application scenarios in the field of autonomous driving. Additionally, we introduce a new Multi-Head Attention Augmented Seq2Seq LSTM Model for vehicle speed prediction, verify its advantage compared with traditional prediction models, and investigate the impact of various input features on model prediction error. 

\item We propose a novel FL framework, FedPAW, which conducts personalized aggregation on the server to capture client-specific information. FedPAW sends tailored aggregated models to clients for personalization, without incurring extra computational and communication overhead to the clients.

\item To the best of our knowledge, we are the first to utilize
PFL for personalized vehicle speed prediction. The effectiveness of our mechanism is validated through experiments, and FedPAW ranks lowest in prediction error within the time horizon of 10 seconds, with a 0.8\% reduction in test MAE, compared to eleven representative benchmark baselines.

\end{itemize}

The remained paper are organized as following. Section \ref{sec:related} introduces the related work. Section \ref{sec:preliminaries} discusses data preparation and prediction models. Section \ref{sec:methodology} presents our proposed framework. The experimental results are described in Section \ref{sec:experiments}. Section \ref{sec:conclusion} serves as the conclusion.

\section{RELATED WORK}
\label{sec:related}
In this section, we investigate the related works of personalized federated learning and vehicle speed prediction, which are the two most concerned research fields in this paper.  The state-of-art advancements are discussed, while their drawbacks when applied in our topic are also analyzed.

\subsection{Personalized Federated Learning}

The traditional FL method, such as FedAvg \cite{mcmahan2017communication}, aims to learn a single global model by aggregating local models from all clients, expecting this model to perform well on most clients. However, this approach often suffers in statistically heterogeneous environments, such as when facing non-IID data, potentially leading to degraded model performance \cite{kairouz2021advances}. FedProx \cite{li2020federated} enhances the stability of the FL process by introducing a proximal term in the client optimization process. However, due to the statistical heterogeneity in FL, obtaining a single global model that fits well across diverse clients remains challenging \cite{kairouz2021advances, huang2021personalized}.

Personalized approaches have gained widespread attention for addressing statistical heterogeneity in FL \cite{kairouz2021advances}. We categorize the PFL methods for aggregating models on the server into the following four categories:

(1) \textit{Meta-learning and fine-tuning}. Per-FedAvg \cite{fallah2020personalized} and FedMeta \cite{chen2018federated} incorporate a meta-learning framework, utilizing the update trend of the aggregated model to learn a global model, easily adapted to local datasets through a few local fine-tuning steps, to achieve good performance on each client.

(2) \textit{Personalized heads}. FedPer \cite{arivazhagan2019federated} divides the neural network model into base and personalized layers, with the base layer designed to be shared across all clients to learn generic feature representations, while the personalized layer remains local to each client for customization. FedRep \cite{collins2021exploiting} similarly learns shared data representations across clients and unique local heads for each client.

(3) \textit{Learning additional personalized models}. pFedMe \cite{t2020personalized} uses Moreau envelopes as regularized loss functions for clients, learning additional personalized models for each client. Ditto \cite{li2021ditto} allows each client to learn its additional personalized model through a proximal term, leveraging information from the global model downloaded from the server.

(4) \textit{Personalized aggregation learning local models on clients}. Recent methods attempt to generate client-specific local models through personalized aggregation on clients \cite{zhang2023fedala}. APFL \cite{deng2020adaptive} obtains personalized models through the aggregation of global and local models, introducing an adaptive learning mixture parameter for each client to control the weight of global and local models. FedFomo \cite{zhang2020personalized} locally aggregates other relevant client models for local initialization in each iteration, effectively computing the best weighted model combination for each client based on how much one client can benefit from another's model. FedALA \cite{zhang2023fedala} captures the necessary information for client models in personalized FL in the global model, with the key component being the Adaptive Local Aggregation (ALA) module, which adaptively aggregates the downloaded global model and local model on clients, thus initializing local models before each iteration of training.

However, these PFL methods require additional personalization steps to be implemented on the client, which typically increases the computational or communication overhead for clients, and may not be appropriate for real-time vehicle speed prediction.

\subsection{Vehicle Speed Prediction}

The trivial methods for vehicle speed prediction assume constant vehicle speed or acceleration \cite{kural2014traffic, han2018safe}. Traditional prediction methods also include Markov Chains \cite{bichi2010stochastic, zhang2011predictive}, Bayesian Networks \cite{kumagai2006prediction, wang2014modeling}. Recurrent Neural Networks (RNN) have proven to be more effective in handling time-series data, hence exhibiting better performance in speed prediction. Some prior studies rely solely on historical speed to predict future velocities, where improvements are sought through enhancing the model structure. Shih et al. \cite{shih2019vehicle} propose a network consisting of encoder-decoder, LSTM, and attention models for predicting vehicle speed. Li et al. \cite{li2023short} introduce a hybrid prediction model that combines K-means, BiLSTM, and GRU.

However, vehicle speed is determined by many complex factors, such as the preceding vehicle and traffic signals. To enhance the accuracy of speed predictors, utilizing information from vehicle-to-vehicle (V2V) and vehicle-to-infrastructure (V2I) communications is crucial \cite{moser2015short}. Han et al. \cite{han2019short} combine a one-dimensional CNN with a BiLSTM to predict vehicle speeds using the information provided by V2V and V2I communications. Jia et al. \cite{jia2020lstm} develope an LSTM-based energy-optimal adaptive cruise control for vehicle speed prediction in urban environments. Wegener et al. \cite{wegener2021longitudinal} explore longitudinal speed prediction in urban settings using V2V and V2I communication information through Conditional Linear Gaussian (CLG) models and LSTM. 
Chada et al. \cite{chada2023deep} conduct speed predictions in urban and highway scenarios using speed limits as well as V2V and V2I communication information, achieving higher prediction accuracy on a simulated driving dataset with LSTM.

Existing vehicle speed prediction models do not consider differences in drivers' driving styles and vehicle types, failing to achieve personalized speed prediction. Moreover, if centralized training is adopted, it does not protect the privacy of drivers' driving data.

\section{Dataset and Prediction Model}
\label{sec:preliminaries}
In this section, we discuss selectino of feature groups and the generation of the approporiate dataset for achieving the training of PFL model to realize personalized vehicle speed prediction. A new LSTM-based Seq2Seq model is also proposed to better predict the vehicle speed by deploying multi-head attention mechanism.

\subsection{Dataset Collection}

Selecting optimized feature parameters can significantly enhance the predictive performance of the model. Thus, it is crucial to carefully select feature parameters closely related to the speed behavior of the target vehicle. 
When traditional methods address the driving in urban scenarios, the prediction of target vehicle's speed mainly deals with the preceding vehicle in the same lane, as well as traffic lights. 
V2V and V2I communications provide additional input from vehicles ahead, while future signal phase and timing (SPaT) information further refine speed predictions \cite{chada2023deep}. 
Our model considers a two-lane context and addresses a more realistic multi-lane environment, since vehicles in adjacent lanes (side vehicles) may also influence the speed of the target vehicle, 
Besides of external information, internal historical control information of the target vehicle (i.e. throttle, brake, and steering wheel angle) directly influences its speed. Furthermore, visual information captured by onboard cameras of the target vehicle, especially processed after semantic segmentation, provides key information reflecting the traffic situation ahead and helping more accurate speed prediction.

\begin{figure}[htbp]
    \centering
    
    \begin{subfigure}[t]{0.25\textwidth}
        \centering
        \includegraphics[width=\textwidth]{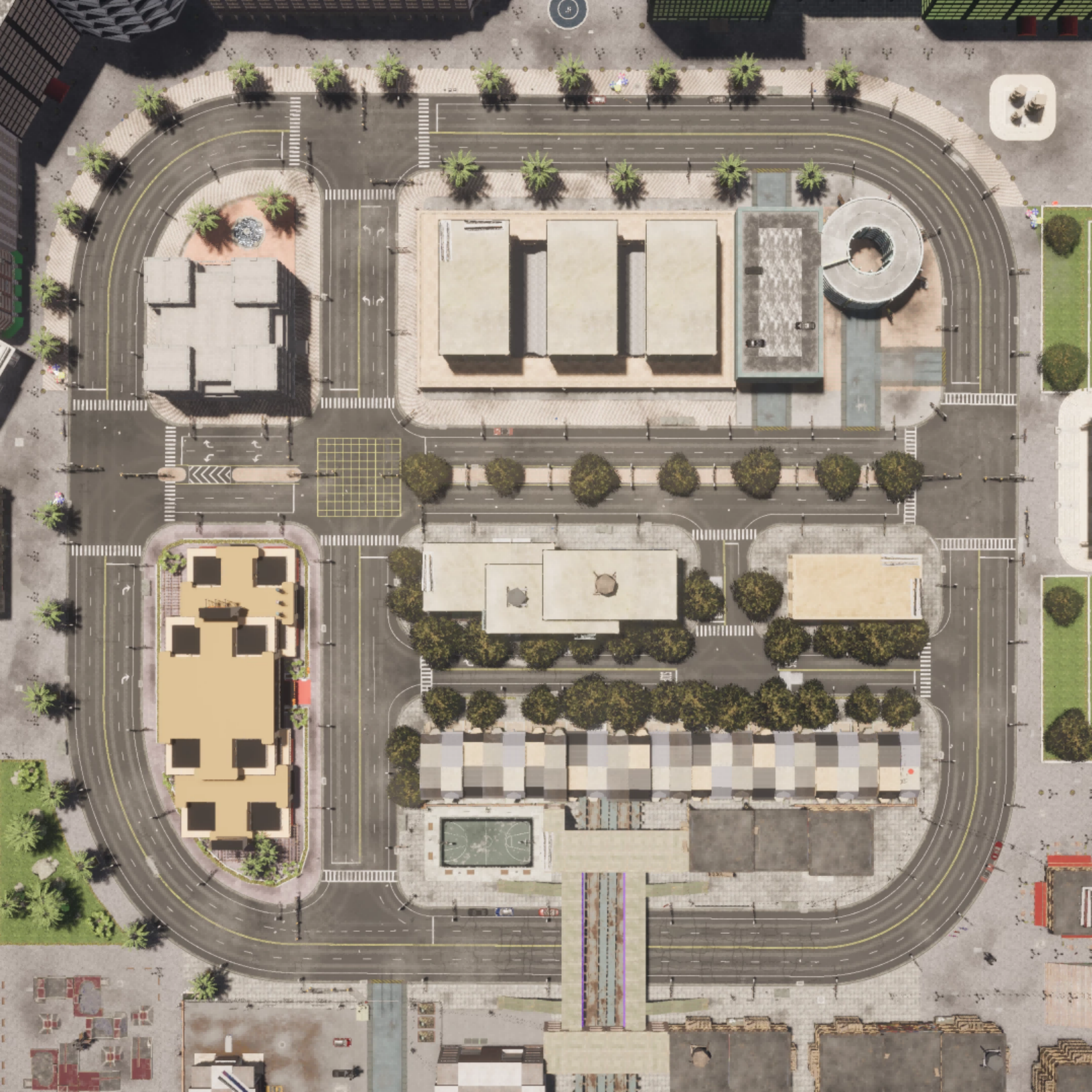}
        \caption{Town 10 in Carla}
        \label{fig:town10_map}
    \end{subfigure}
    \hfill
    \begin{minipage}[t]{0.2\textwidth}
        \centering
         \vspace{-128pt}
        \begin{subfigure}[t]{\textwidth}
            \centering
            \includegraphics[width=\textwidth]{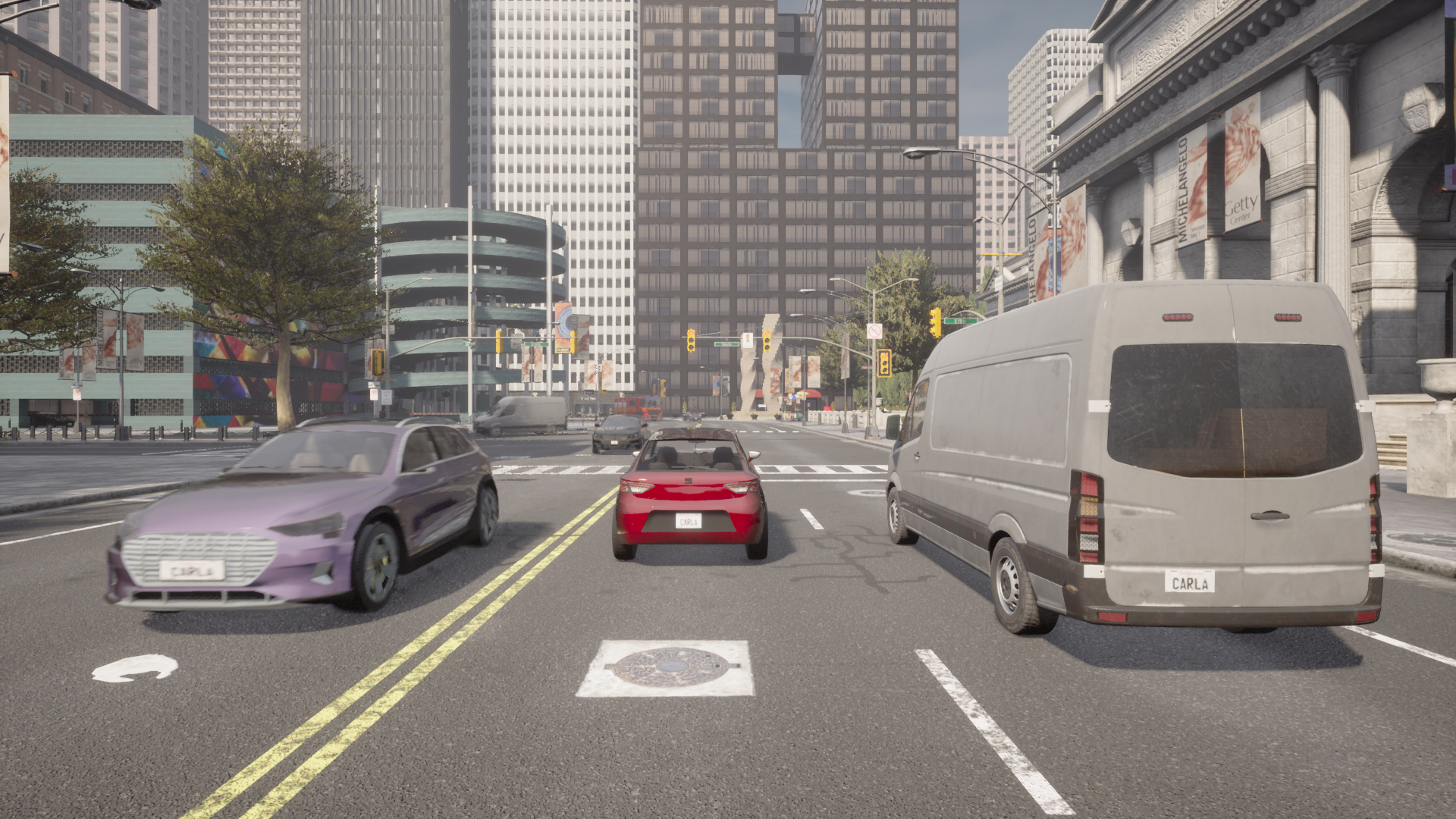}
            \caption{RGB image in Carla}
            \label{fig:rgb_image}
        \end{subfigure}
        
        
        \begin{subfigure}[t]{\textwidth}
            \centering
            \includegraphics[width=\textwidth]{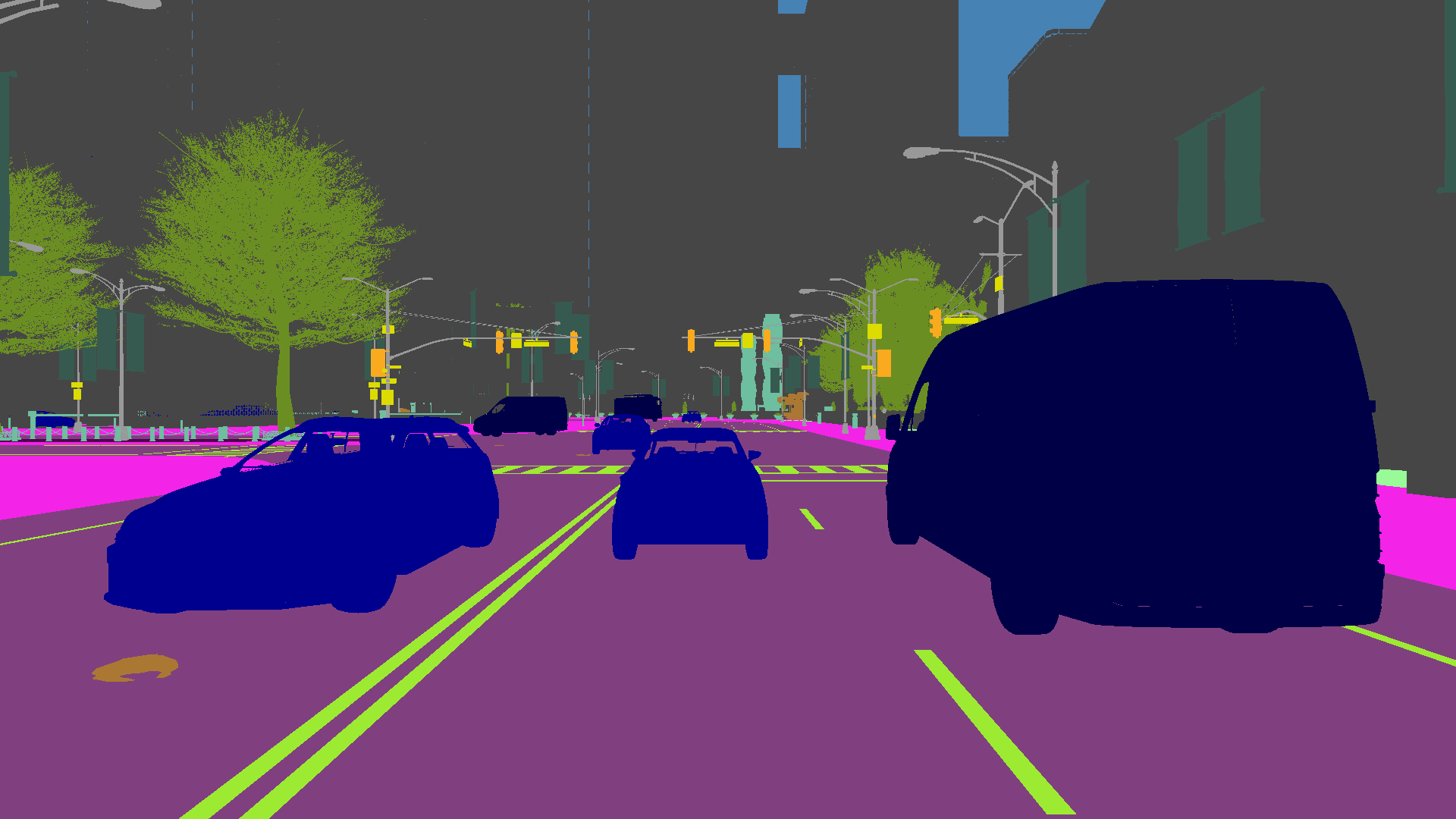}
            \caption{Semantic segmentation image}
            \label{fig:semantic_segmentation_image}
        \end{subfigure}
    \end{minipage}
    
    \caption{Data collection scenarios. (a): Urban map Town 10 in Carla simulator. (b) and (c): RGB image and Semantic segmentation image captured by the camera sensor in Carla simulator.}
    \label{fig:carla}
\end{figure}

Unlike other data preparation methods that consider fixed drivers, vehicles, and driving routes \cite{jia2020lstm, wegener2021longitudinal}, our study collects driving data through manual driving of various  types of vehicles by different drivers in the Carla simulation driving environment. Our CarlaVSP dataset\footnote{CarlaVSP dataset is open sourced at: https://pan.baidu.com
/s/1qs8fxUvSPERV3C9i6pfUIw?pwd=tl3e.} is independent of driving routes but related to drivers and vehicles. Extracting V2V and V2I information in the real world is both laborious and highly expensive, and there is a lack of public driving datasets distinguishing between drivers and vehicles. Using a Logitech G29 steering wheel in the city center scenario map of the Carla simulator (Town 10, as shown in Figure \ref{fig:town10_map}), we assigned ten volunteers to drive three different types of vehicles manually, containing Nissan Micra, Audi A2, and Tesla Model 3, and collected about 1 hour of driving data from each volunteer.

\begin{figure}[htpb]
    \centering
    \includegraphics[width=0.8\linewidth]{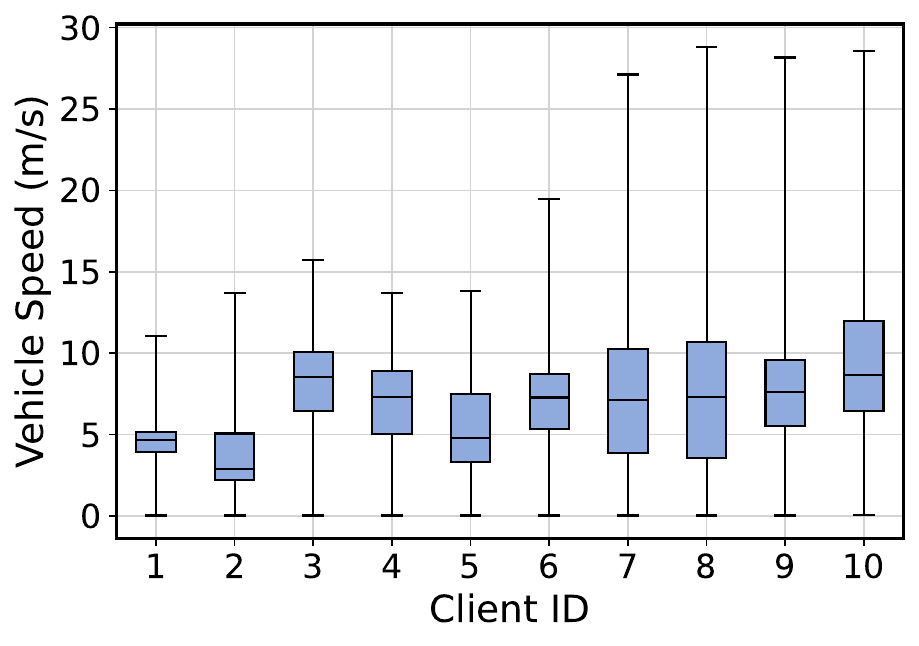}
    \caption{Vehicle speed distribution of 10 clients during driving (excluding parking).}
    \label{fig:velocity_distribution}
\end{figure}

The diverse driving styles of different volunteers and the power differences of various vehicle types provide an ideal environment for simulating non-IID driving data in the real world. During driving (excluding parking), the vehicles driven by these ten volunteers, acting as ten clients, exhibit significant differences in speed distribution, as shown in Figure \ref{fig:velocity_distribution}. Existing driving datasets do not label drivers and vehicle types, whereas our CarlaVSP dataset includes these labels. Furthermore, the CarlaVSP dataset is not only useful for vehicle speed prediction but also for other personalized applications in the field of autonomous driving. Moreover, the target vehicles in the Carla simulator are equipped with camera sensors capable of capturing RGB images of the road ahead (as shown in Figure \ref{fig:rgb_image}) and the corresponding semantic segmentation images (as shown in Figure \ref{fig:semantic_segmentation_image}), allowing traffic elements ahead to be conveniently incorporated into the driving data.

\begin{figure}[!ht]
    \centering
    \includegraphics[width=\linewidth]{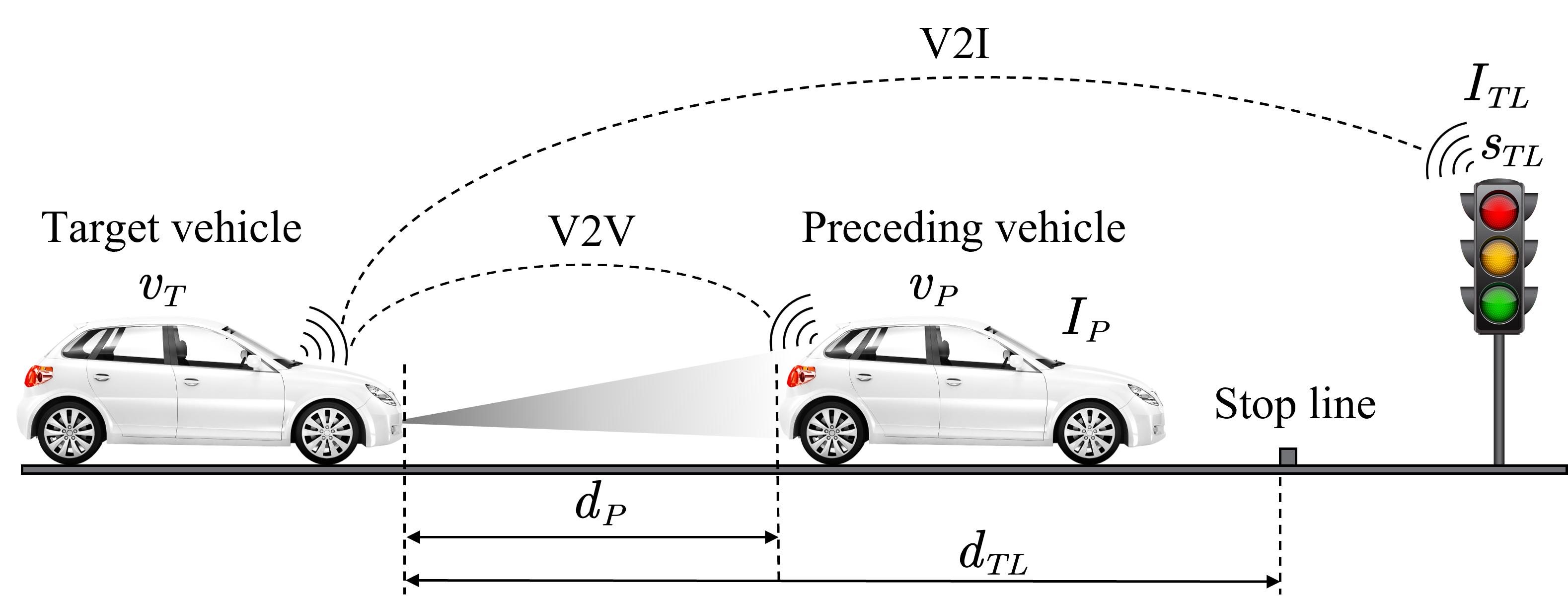}
    \caption{Schematic of V2V and V2I enabled traffic scenario.}
    \label{fig:traffic_scenario}
\end{figure}

We aim to predict the speed scalar of the target vehicle within a specific range in the future for discrete-time settings at 1-second intervals based on some input features. Table \ref{tab:feature_groups} and Figure \ref{fig:traffic_scenario} show seven feature groups named from FG1 to FG7, where $k$ represents the current discrete time index. The most appropriate feature group is required to generate the most accurate vehicle speed prediction. As shown in  Figure \ref{fig:traffic_scenario}, feature group FG1 consists of multiple input features: target vehicle speed $v_T$, speed of the preceding vehicle $v_P$, indicator of preceding vehicle existence $I_P$, indicator of traffic light existence $I_{TL}$, the distance to the preceding vehicle $d_P$, the distance to the traffic light stop line $d_{TL}$, current traffic light signal state $s_{TL}$, and future traffic light states $s_{TL,k+1},...,s_{TL,k+H}$. Here, $H$ is the prediction time horizon. Some settings of the features are given, e.g. if the distance to the preceding vehicle or traffic light exceeds 100 meters or they are not present, then $I_P$ and $I_{TL}$ are set to False; otherwise, set to True. FG2, based on FG1, further considers side vehicles, adding three input features: the side vehicle speed $v_S$, indicator of  a side vehicle existence $I_S$, and the distance to the side vehicle $d_S$. FG3, based on FG1, adds historical control information of the target vehicle: throttle amplitude $throttle$, brake amplitude $brake$, and steering wheel angle $steer$. FG4 adds traffic element proportion in image (TEPI) on top of FG1 to assess the traffic situation ahead. Specifically, TEPI provides an intuitive indicator of the traffic condition ahead by calculating the proportion of cars and vans $r_1$, buses and trucks $r_2$, and traffic lights $r_3$ in the total pixels of the image. These ratios reflect the visual proportion of different traffic elements on the road. Feature Groups FG5, FG6, and FG7 are combinatorial unions of FG2 to FG4,
aiming to explore the impact of different feature combinations.

\begin{table}[!ht]
  \centering
  \caption{Feature groups and corresponding input feature sets}
  \begin{tabular}{cl}
    \toprule
    Groups & Input Feature Sets \\
    \midrule
    FG1 & \{$v_T,v_P,I_P,I_{TL},d_P,d_{TL},s_{TL},s_{TL,k+1},...,s_{TL,k+H}$\} \\
    FG2 & FG1 $\cup$ \{$v_S,I_S,d_S$\} \\
    FG3 & FG1 $\cup$ \{$throttle,brake,steer$\} \\
    FG4 & FG1 $\cup$ \{$r_1,r_2,r_3$\} \\
    FG5 & FG2 $\cup$ FG4 \\
    FG6 & FG3 $\cup$ FG4 \\
    FG7 & FG2 $\cup$ FG3 $\cup$ FG4 \\
    \bottomrule
  \end{tabular}
  \label{tab:feature_groups}
\end{table}




\subsection{Multi-Head Attention Augmented Seq2Seq LSTM Model}

\begin{figure}[htpb]
    \centering
    \includegraphics[width=\linewidth]{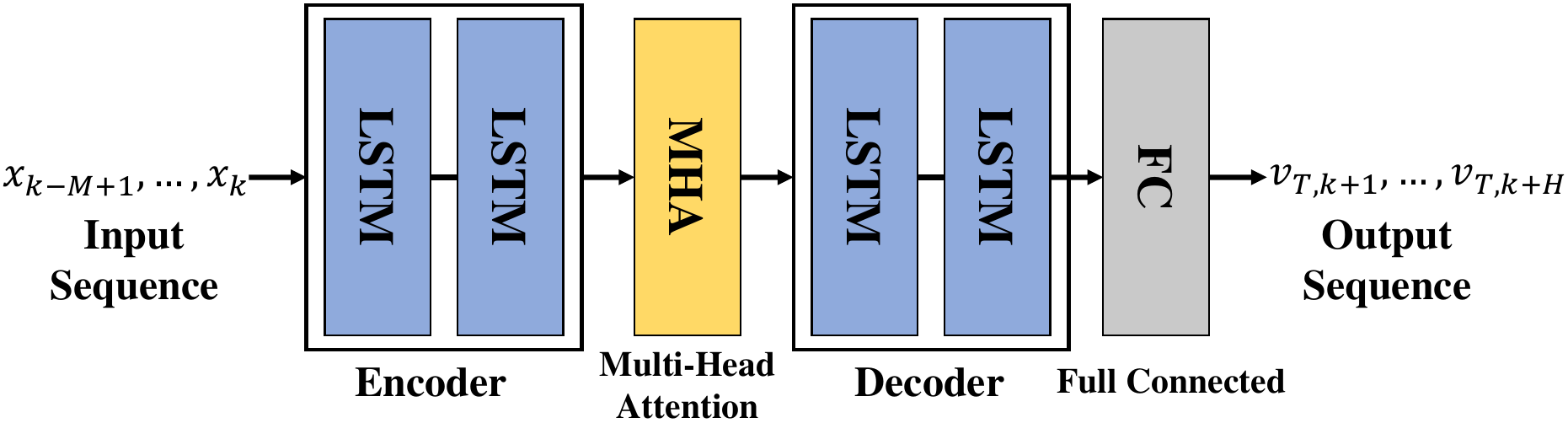}
    \caption{Multi-Head Attention Augmented Seq2Seq LSTM Model.}
    \label{fig:seq2seq_lstm}
\end{figure}

As shown in Figure \ref{fig:seq2seq_lstm}, we propose the Multi-Head Attention Augmented Seq2Seq LSTM Model for vehicle speed prediction, which forecasts the future speed over a period based on a segment of historical feature data. The feature groups are organized in chronological order into an input sequence $x_{k-M+1},...,x_{k}$, with the future speed of the target vehicle as the output sequence $v_{T,k+1},...,v_{T,k+H}$, where $k$ represents the current discrete time index, $M$ denotes the past time range, and $H$ represents the forecasting horizon. The Seq2Seq architecture, originated from the domain of machine translation \cite{cho2014learning}, has been proven effective in various sequence modeling tasks due to its encoder-decoder structure. Our model further incorporates Long Short-Term Memory (LSTM) units that are capable of capturing temporal dependencies, addressing challenges posed by the sequential nature of the vehicle speed prediction task. 

The integration of the Multi-Head Attention mechanism, inspired by the transformer model architecture \cite{vaswani2017attention}, enables the model to capture complex temporal dynamics and long-distance dependencies within sequences. Specifically, the model processes the input sequence through a two-layer LSTM encoder, extracting a high-dimensional representation of temporal data. This is followed by a custom Multi-Head Attention module, which captures dependencies within the sequence across multiple subspaces in parallel. The output from this attention mechanism is then fed into a two-layer LSTM decoder, which incrementally constructs predictions for future vehicle speeds. Finally, a fully connected layer maps the LSTM output to the predicted speed. Our model combines LSTM capability for processing sequential data and multi-head attention mechanism advantage in capturing long-term dependencies, aiming to enhance speed prediction accuracy.

\section{METHODOLOGY}
\label{sec:methodology}
In this section, we first provide an overview of the proposed FedPAW framework, then state the objectives of FL optimization, and finally depict the formal derivation of FedPAW. For ease of reading, the main notations are listed in Table \ref{tab:notations}.

\begin{table}[!ht]
  \centering
  \caption{Summary of Main Notations}
  \begin{tabular}{c p{0.78\linewidth}}
    \toprule
    Notation & Description \\
    \midrule
    $N$ & The number of clients  \\
    $D_i$ & The local dataset of client $i$ \\
    $\hat{\Theta}_i$ & Aggregated model of client $i$ \\
    $\mathcal{L}_i$ & Local loss function of client $i$\\
    $\mathcal{G}$ & Global loss function \\
    $k_i$ & The contribution of client $i$ to the dataset\\
    $t$ & Iteration index \\
    $S^t$ & The subset of $N$ clients for training at $t$-th iteration\\
    $\Theta _{i}^t$ & Local model of client $i$ at $t$th iteration \\
    $\Theta ^{t}$ & Global model at $t$th iteration \\
    $\hat{\Theta}_{i}^{t+1}$ & Aggregated model of client $i$ at $(t+1)$-th iteration \\
    $\odot$ & Hadamard product\\
    $W^{t}$ & Aggregation weights at $t$-th iteration \\
    $p$ & The range of personalized aggregation \\
    $L(\Theta_i)$ & The layer count in $\Theta ^{t}_i$ \\
    $r$ & The number of iterations before personalized aggregation \\
    $W^{t,p}$ & Weighted model parameter difference measure at $t$-th iteration \\
    $\rho$ & Client joining ratio \\ 
    $\eta$ & Local learning rate \\

    \bottomrule
  \end{tabular}
  \label{tab:notations}
\end{table}

\subsection{Overview of FedPAW}
\begin{figure*}[htpb]
    \centering
    \includegraphics[width=\textwidth]{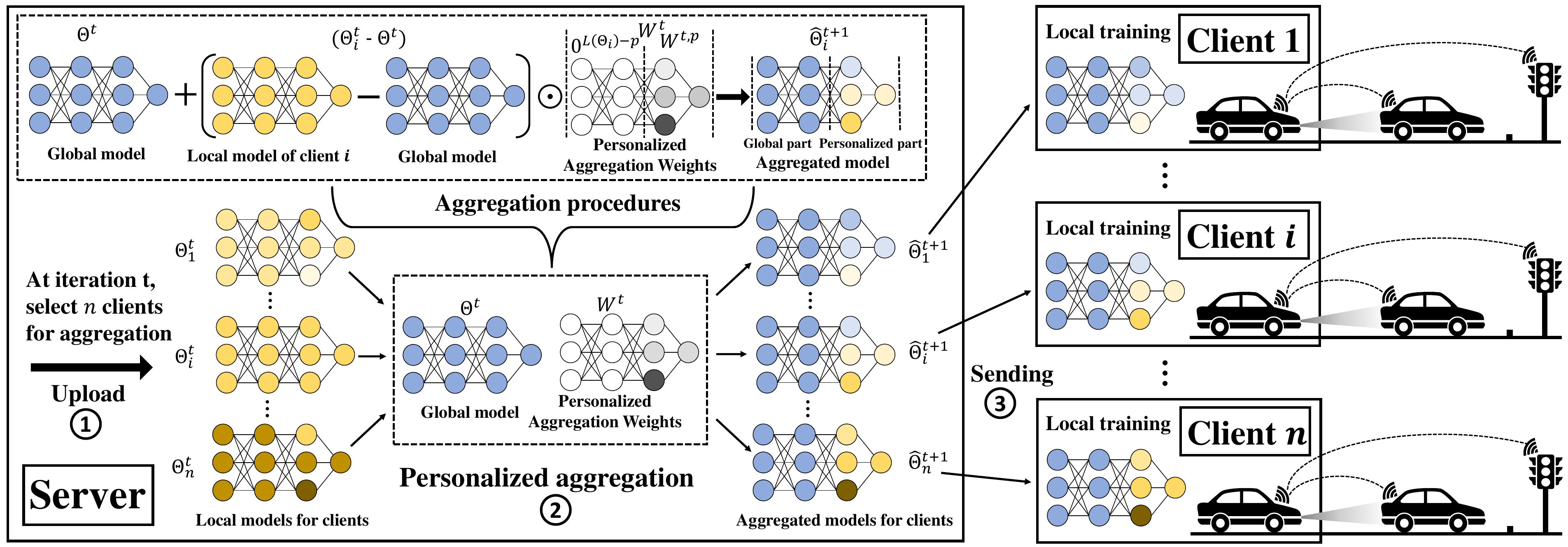}
    \caption{Overview of the FedPAW Framework: \ding{172} Clients upload local models to the server. \ding{173} The server performs personalized aggregation using the client local models and the global model to produce the aggregated models. \ding{174} The server sends tailored aggregated models to the clients. The figure uses blue color to denote global model parameters, yellow color to denote local model parameters, and grey color to denote aggregated weights. Darker colors reflect higher values.}
    
    
    \label{fig:fedpaw}
\end{figure*}

Figure \ref{fig:fedpaw} illustrates the proposed Federated Learning with Personalized Aggregation Weights (FedPAW) framework, which implements a personalized aggregation (PA) step on the server. In each iteration, the server initially selects some clients for PA, where clients upload trained local models to the server. The server then performs an average aggregation of these client local models to obtain a global model,
followed by an additional PA step. Specifically, the server element-wisely aggregates the client local models and the global model to produce aggregated models. Finally, the server distributes tailored aggregated models to each client, facilitating personalized speed prediction. Note that, compared to FedAvg, the proposed FedPAW only adds some basic matrix operations, the computation cost of which could be neglected in the server.

\subsection{Problem Statement}

In the FL process, multiple target vehicles act as clients, with their learning tasks coordinated by a central server. Suppose there are $N$ clients, each possessing their own private training datasets $D_1,...,D_N$, and these datasets are Non-IID. Under the coordination of the central server, the overall goal is to collaboratively learn independent local models $\hat{\Theta}_1,...,\hat{\Theta}_N$ for each client 
without uploading their private data. The aim is to minimize the global loss function and obtain reasonable local models:
\begin{equation}
\{\hat{\Theta}_1,...,\hat{\Theta}_N\}=\argmin_{\hat{\Theta}}
\mathcal{G}(\mathcal{L}_1,...,\mathcal{L}_N)
\end{equation}
where $\mathcal{L}_i=\mathcal{L}(\hat{\Theta}_i,D_i), \forall i\in [1,N]$, and $\mathcal{L}\left( \cdot \right)$ is the loss function, representing the mean squared error (MSE) between the predicted vehicle speeds and the actual values. Typically, $\mathcal{G}\left( \cdot \right)$ is set to $\sum_{i=1}^N{k_i}\mathcal{L}_i$, where $k_i=|D_i|/\sum_{j=1}^N{|}D_j|$ measuring the contribution of client $i$ to the dataset, and $|D_i|$ is the number of local data samples of client $i$.

\subsection{Federated Learning with Personalized Aggregation Weights (FedPAW)}

In traditional FL such as FedAvg, at iteration $t$, the server selects a subset of $N$ clients $S^t$ for training, and aggregates all local models $\Theta _{i}^t,i \in S^t$ to form the global model $\Theta ^{t}$. It is obtained as follows:

\begin{equation}
\Theta^t \leftarrow \sum_{i \in S^t} \frac{k_i}{\sum_{j \in S^t} k_j} \Theta_i^t
\label{eq:get_global_model}
\end{equation}

At iteration $t+1$, after the server sends the global model $\Theta ^{t}$ to client $i$, $\Theta ^{t}$ overrides the previous local model $\Theta _{i}^{t}$, resulting in the local model for this round $\hat{\Theta}_{i}^{t+1}$ for local training, i.e., $\hat{\Theta}_{i}^{t+1}:=\Theta ^{t}$. In FedPAW, the server element-wisely aggregates the global model and the client local models, then sends the aggregated model $\hat{\Theta}_{i}^{t+1}$, not $\Theta ^{t}$, to client $i$. Formally:

\begin{equation}
\begin{aligned}
\hat{\Theta}_{i}^{t+1}:=\Theta ^{t}\odot W_{1}^{t}+\Theta _{i}^{t}\odot W_{2}^{t}\\
s.t.  w_{1}^{q}+w_{2}^{q}=1,\forall \,\mathrm{valid}\,q.
\end{aligned}
\label{eq:get_p_model1}
\end{equation}

Where $\odot$ is the Hadamard product, representing element-wise multiplication of two matrices, $w_{1}^{q}$ and $w_{2}^{q}$ are the $q$-th parameters of aggregation weights $W_{1}^{t}$ and $W_{2}^{t}$, respectively. 
FedAvg is a special case of the proposed FedPAW, where
$\forall \,\mathrm{valid}\,q, w_{1}^{q}\equiv 1$ and $w_{2}^{q}\equiv 0$. 

The two weights $W_{1}^{t}$ and $W_{2}^{t}$, with the above constraint, can be further simplified in form. We reformulate Eq. (\ref{eq:get_p_model1}) by merging $W_{1}^{t}$ and $W_{2}^{t}$ into:

\begin{equation}
\hat{\Theta}_{i}^{t+1}:=\Theta ^{t}+(\Theta _{i}^{t}-\Theta^{t})\odot W^{t},
\label{eq:get_p_model2}
\end{equation}
where $w \in [0,1],\, \forall w \in W^{t}$.

Since lower layers of DNN learn more general information than higher layers \cite{zhu2021data}, clients could obtain most general information from the lower layers of the global model \cite{zhang2023fedala}. To reduce computational overhead and improve model performance,  a hyperparameter $p$ is introduced to control the scope of PA, which is applied to the top $p$ layers, and the global model $\Theta ^{t}$ is used to override the remained lower layers just like FedAvg:

\begin{equation}
\hat{\Theta}_{i}^{t+1}:=\Theta ^{t}+(\Theta _{i}^{t}-\Theta ^{t})\odot [0^{L(\Theta_i)-p};W^{t,p}]
\label{eq:get_p_model3}
\end{equation}

Where $L(\Theta_i)$ is the layer count of $\Theta ^{t}_i$, and $0^{L(\Theta_i)-p}$ matches the shape of the lower layers of $\Theta ^{t}_i$, 
with elements as zeros. The aggregation weights $W^{t,p}$ match the shape of the top $p$ layers.

The design of the aggregation weights $W^{t,p}$ focuses on evaluating the local model parameters that can provide significant personalized information to the aggregated model $\hat{\Theta}_{i}^{t+1}$. Theoretically, we expect those local model parameters offering significant personalized information to receive more weights in $W^{t,p}$. In practice, identifying these "personalized model parameters" is challenging. However, we can make an empirical assumption: typically, after multiple rounds of FedAvg, parameters of different client local models should converge closely, including the parameters of the average aggregated global model. However, due to the non-IID of clients' data, significant differences in specific parameters among different client local models indicate that these parameters actually contain a wealth of personalized information.

Thus, increasing the weight of these parameters during PA can capture client-specific personalized information and enhance the predictive capability of aggregated models.

To ensure that client local models undergo sufficient FedAvg rounds before PA takes place, another hyperparameter $r$ is considered. For iterations $t < r$, each element in $W^{t,p}$ is set to zero, making the algorithm identical to FedAvg at this stage. Upon reaching iteration $t \geq r$, the server performs an adaptive calculation of the personalized aggregation weights $W^{t,p}$. Specifically, we calculate the weighted mean squared error between all local models of client subset $S^t$ and the global model for the top $p$ layers (denoted as $\Theta^{t,p}_i, i \in S^t$ and $\Theta ^{t,p}$), to obtain the Weighted Model Parameter Difference Measure $M^{t,p}$, which reflects the degree of model parameter differences between different clients, weighted by the number of local data samples. Formally:

\begin{equation}
M^{t,p} \leftarrow \sum_{i \in S^t} \frac{k_i}{\sum_{j \in S^t} k_j} (\Theta_i^{t,p}-\Theta^{t,p})\odot(\Theta_i^{t,p}-\Theta^{t,p})
\label{eq:get_m}
\end{equation}

Each layer of $M^{t,p}$, denoted by $M^{t,p_l}$, is normalized to derive the aggregation weights $W^{t,p}$, ensuring that $w \in [0,1],\, \forall w \in W^{t,p}$. Formally:

\begin{equation}
W^{t,p} \leftarrow \frac{M^{t,p_l}-min(M^{t,p_l})}{max(M^{t,p_l})-min(M^{t,p_l})}, \forall l\in [1,p]
\label{eq:get_w}
\end{equation}

\begin{figure}[htpb]
    \centering
    \includegraphics[width=\linewidth]{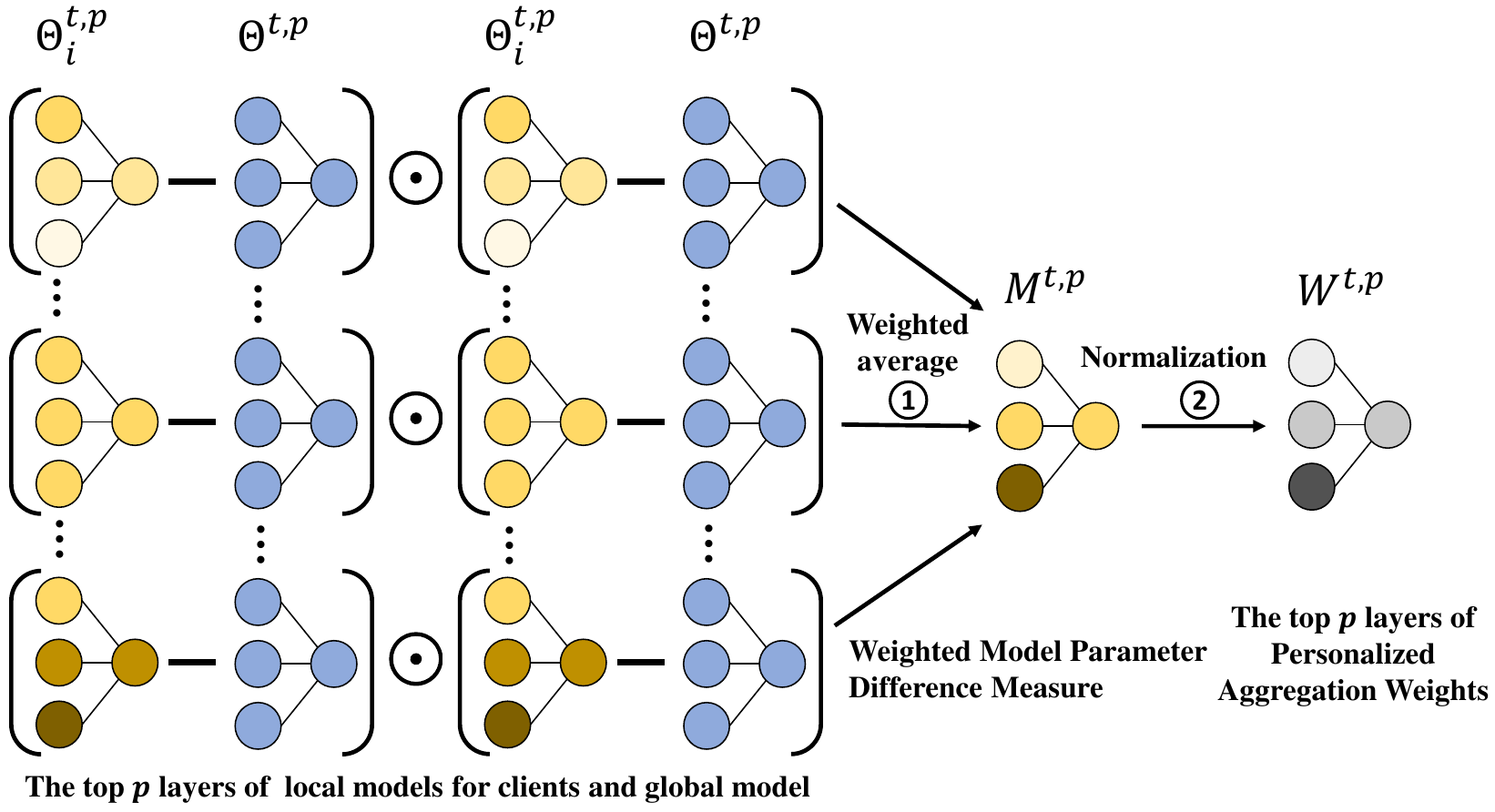}
    \caption{The computation process for the top $p$ layers of Personalized Aggregation Weights involves: \ding{172} Calculating the weighted mean squared error between the top $p$ layers of client local models and global model. \ding{173} Normalizing the calculated Weighted Model Parameter Difference Measure.}
    \label{fig:get_paw}
\end{figure}

Figure \ref{fig:get_paw} illustrates the computation process for the top $p$ layers of Personalized Aggregation Weights. Algorithm \ref{alg:fedpaw} describes the entire FL process within FedPAW.

\begin{algorithm}[t!]
\caption{FedPAW}
\label{alg:fedpaw}
\KwIn{$N$ clients, client joining ratio $\rho$, loss function $\mathcal{L}$, initial global model $\Theta^0$, local learning rate $\eta$, the number of iterations $r$ to wait, the range of PA $p$.}
\KwOut{Reasonable local models $\hat{\Theta}_1, \ldots, \hat{\Theta}_N$}

Server sends $\Theta^0$ to all clients to initialize local models.\\
\For {\textnormal{iteration} $t = 1, \ldots, T$}
{Server samples a subset $S^t$ of clients based on $\rho$.\\
Server sends $\hat{\Theta}^t_i$ to all client $i \in S^t$.\\
    \For {\textnormal{Client} $i \in S^t$ \textnormal{in parallel}}{
        Client $i$ obtains $\Theta^t_i$ by  \hfill $\vartriangleright$ \textbf{Local model training} \\
         \Indp
         $\Theta_i^t \leftarrow \hat{\Theta}_i^t - \eta \nabla_{\hat{\Theta}_i} {\mathcal{L}}(\hat{\Theta}_i^t, D_i)$\\
         \Indm
         Client $i$ sends $\Theta_i^t$ to the server. \hfill $\vartriangleright$ \textbf{Uploading}
    }

Server obtains $\Theta^t$ by Eq. (\ref{eq:get_global_model}).\\
\eIf{$t < r$}{
Server sets elements of $W^{t,p}$ to 0. \hfill $\vartriangleright$ \textbf{FedAvg}
} 
{
Server obtains $W^{t,p}$ by Eq. (\ref{eq:get_m}) and (\ref{eq:get_w}). \hfill $\vartriangleright$ \textbf{PA}
}
Server obtains $\hat{\Theta}^{t+1}_i$ for client $i \in S^t$ by Eq. (\ref{eq:get_p_model3}).\\
}
\textbf{return} $\hat{\Theta}_1, \ldots, \hat{\Theta}_N$

\end{algorithm}

\section{EXPERIMENTS}
\label{sec:experiments}
\subsection{Experimental Setup}
In this section, extensive experiments are conducted to assess the performance of the FedPAW framework. Initially, we explore the impact of different feature groups (FG1 to FG7) on the prediction model. The learning methods are centralized training based on cloud servers (Cloud) and local training (Local). Additionally, the comparisons include the physics-based Constant Velocity (CV) and Constant Acceleration (CA) models, as well as traditional LSTM models. Then, we investigate the effects of the hyperparameters $r$ and $p$ of the FedPAW. Subsequently, based on the selected feature groups and hyperparameters, a comprehensive comparison is conducted between FedPAW and eleven benchmark baselines, including Cloud, Local, FedAvg \cite{mcmahan2017communication}, FedProx \cite{li2020federated} and seven SOTA PFL methods, namely: Per-FedAvg \cite{fallah2020personalized}, FedRep \cite{collins2021exploiting}, pFedMe \cite{t2020personalized}, Ditto \cite{li2021ditto}, APFL \cite{deng2020adaptive}, FedFomo \cite{zhang2020personalized}, and FedALA \cite{zhang2023fedala}. Finally, further analysis is performed on the speed prediction results of the prediction model.

The dataset used is the 10-hours simulated driving data CarlaVSP that we collected in Section 3.1, distributed across ten clients, with their speed distributions being Non-IID, as shown in Figure \ref{fig:velocity_distribution}. In all the experiments, for the 5-second (5-s) prediction, the encoder and decoder of the prediction model utilize two LSTM layers with a dropout rate of 0.1 to reduce overfitting, while for the 10-second (10-s) prediction, three LSTM layers are used with a dropout rate of 0.2. The traditional LSTM model used for comparison utilizes two LSTM layers with a dropout rate of 0.1. The hidden dimension of the LSTM layer is consistently set to 128. The length of the input history sequence for the prediction model is equal to that of the output prediction sequence. Moreover, Mean Square Error (MSE) is used as the loss function, with the ADAM \cite{kingma2014adam} optimizer and a local learning rate set to 0.005. The local batch size is 64, the number of local model training epochs is 1, and all FL methods are run for 300 iterations to ensure empirical convergence. Correspondingly, the Cloud and Local methods undergo 50 and 200 iterations, respectively, to achieve empirical convergence.

The evaluation metrics for the prediction model are the Mean Absolute Error (MAE) and Root Mean Square Error (RMSE) between the predicted and actual speeds. For traditional FL, the global model with the lowest test error is selected; for PFL, the local models with the lowest average test errors are chosen. All prediction models are evaluated on the clients, with 20\% of the local data used as the test dataset and the remaining 80\% used for training. All methods were run five times, and the results are presented as the average and standard deviation of the evaluation metrics.

In the experiments, we implement FedPAW using PyTorch-1.12.1 and simulate FL on a server equipped with an AMD Epyc 7302 16-core processor x 64, 8 NVIDIA GeForce RTX 3090 GPUs, 472.2GB of memory, and running Ubuntu 20.04.5 operating system.

\begin{table*}[htbp] 
\centering 
\caption{The impact of different feature groups on the prediction error (m/s) of the model.}
\begin{tabular}{ll|lllllll|lllllll}
\toprule
\multicolumn{2}{l|}{Horizons}   & \multicolumn{7}{c|}{5s}                                                     & \multicolumn{7}{c}{10s}                                        \\
\midrule
\multicolumn{2}{l|}{Feature Groups}   & \multicolumn{1}{c}{FG1} & \multicolumn{1}{c}{FG2} & \multicolumn{1}{c}{FG3} & \multicolumn{1}{c}{FG4} & \multicolumn{1}{c}{FG5} & \multicolumn{1}{c}{\textbf{FG6}} & \multicolumn{1}{c|}{FG7} & \multicolumn{1}{c}{FG1} & \multicolumn{1}{c}{FG2} & \multicolumn{1}{c}{FG3} & \multicolumn{1}{c}{FG4} & \multicolumn{1}{c}{FG5} & \multicolumn{1}{c}{\textbf{FG6}} & \multicolumn{1}{c}{FG7} \\
\midrule
\multirow{2}{*}{MAE}     & Cloud   & 1.216 & 1.217 & 1.165 & 1.211 & 1.250 & \textbf{1.161} & 1.171 & 1.701 & 1.731 & 1.699 & 1.717 & 1.739 & \textbf{1.695} & 1.713 \\
                         & Local   & 1.258 & 1.309 & 1.184 & 1.287 & 1.299 & \textbf{1.176} & 1.228 & 1.800 & 1.840 & 1.724 & 1.815 & 1.859 & \textbf{1.721} & 1.739 \\
\midrule
\multirow{2}{*}{RMSE}    & Cloud   & 1.437 & 1.437 & 1.391 & 1.432 & 1.470 & \textbf{1.386} & 1.397 & 2.079 & 2.115 & 2.101 & 2.108 & 2.118 & \textbf{2.077} & 2.089 \\
                         & Local   & 1.479 & 1.536 & 1.409 & 1.502 & 1.523 & \textbf{1.395} & 1.451 & 2.177 & 2.230 & 2.107 & 2.214 & 2.245 & \textbf{2.099} & 2.126\\
\bottomrule
\end{tabular}
\label{tab:feature_groups_test}
\end{table*}

\begin{table*}[htbp] 
\centering 
\caption{The impact of different hyperparameters on the prediction error (m/s) of FedPAW at a prediction horizon of 5 s.}
\begin{tabular}{l|llllll|lllllll}
\toprule
 & \multicolumn{6}{c|}{$p=2$}                                                 & \multicolumn{7}{c}{$r=1$}                                                           \\ 
\midrule
Items                                & $r=1$     & $r=2$             & $r=3$     & $r=5$     & $r=10$     & $r=20$     & $p=1$      & $p=2$               & $p=3$      & $p=4$     & $p=5$      & $p=6$      & $p=max$    \\
\midrule
MAE   & \textbf{1.166} & 1.170 & 1.179 & 1.188 & 1.183 & 1.266 & 1.173 & \textbf{1.166} & 1.172 & 1.190 & 1.191 & 1.180 & 1.178 \\
\midrule
RMSE  & \textbf{1.379} & 1.382 & 1.385 & 1.404 & 1.397 & 1.481 & 1.381 & \textbf{1.379} & 1.382 & 1.399 & 1.399 & 1.390 & 1.391 \\
\bottomrule
\end{tabular}
\label{tab:hyper}
\end{table*}

\begin{table}[htbp]
\centering 
\caption{The prediction error (m/s) for the CV, CA and traditional LSTM models.}
\begin{tabular}{l|ll|ll}
\toprule
Metrics       & \multicolumn{2}{c|}{MAE} & \multicolumn{2}{c}{RMSE} \\
\midrule
Horizons & 5s          & 10s        & 5s          & 10s        \\
\midrule

CV   & 2.389      & 3.007      & 2.642       & 3.459      \\
CA     & 2.447      & 3.130      & 2.711       & 3.598      \\
LSTM (Cloud+FG6)     & \textbf{1.212} & \textbf{1.768} & \textbf{1.444} & \textbf{2.155} \\
LSTM (Local+FG6)     & 1.232      & 1.789      & 1.456       & 2.171      \\
\bottomrule
\end{tabular}
\label{tab:cvca}
\end{table}

\subsection{Impact of Feature Groups}
The selection of feature parameters is crucial for the performance of the prediction model. Table \ref{tab:feature_groups_test} displays the average MAE and RMSE of the proposed prediction model across seven feature groups (FG1 to FG7) over different prediction horizons (5 s and 10 s), with Cloud and Local learning methods being utilized. For a 5 s prediction horizon, the performance of FG3 is superior to that of FG1, and FG4 shows a slightly better performance than FG1 for Cloud, while FG2 is inferior to FG1. This indicates that the additional features, namely historical control information, have a positive contribution to the performance of the prediction model, TEPI contributes marginally, and side vehicle information has a negative impact, possibly due to the infrequent lane changes of vehicles in the Carla simulator. For a 10 s prediction horizon, as the prediction horizon increases, the reference value of historical feature information decreases, leading to a diminished performance of FG3 and FG4 compared to the earlier. FG5 to FG7 are combinations of these additional features, and for both 5 s and 10 s prediction horizons, the best-performing feature group is FG6. Therefore, we have chosen FG6 (i.e. the union of FG1, the control information, and the traffic element proportions) for subsequent experiments. Compared to Table \ref{tab:cvca}, it is evident that the predictive performance of the proposed model significantly surpasses that of CV, CA, and traditional LSTM models.

\subsection{Impact of Hyperparameters}
The impact of the hyperparameters $r$ and $p$ in the FedPAW framework, as shown in Table \ref{tab:hyper}, is evaluated with a prediction horizon of 5 s. For the hyperparameter $r$, reducing $r$ allows for personalized aggregation to be performed in earlier iteration rounds, thus exhibiting the best performance at $r=1$. Generally, for computationally complex tasks, a $r$ with very low value would prevent sufficient rounds of FedAvg from occurring in advance, resulting in client local models not converging adequately. This could hinder the differentiation of parameters containing more personalized information. However, given that our task is computationally simpler and the models converge quickly, this effect is insignificant. We set $r=1$ for prediction horizons of both 5 s and 10 s. As for the hyperparameter $p$, decreasing $p$ reduces the number of layers involved in PA, making effective use of the general information from the lower layers of the global model, thereby reducing prediction loss. Nonetheless, when $p$ is reduced to 1, the scope of PA becomes too small to fully leverage the advantages of FedPAW, thus $p=2$ shows the best performance. We set $p=2$ for a prediction horizon of 5 s and adjust $p=4$ for the prediction horizon of 10 s based on similar experiments.

\subsection{Performance Comparison and Analysis}

\begin{table*}[htpb] 
\centering 
\caption{The prediction error (m/s) at different prediction horizons and different client joining ratio $\rho$.}

\begin{tabular}{l|llll|llll}
\toprule
Horizons   & \multicolumn{4}{c|}{5s}                                                                      & \multicolumn{4}{c}{10s}                                                                     \\
\midrule
Ratios      & \multicolumn{2}{c}{$\rho=1$}                    & \multicolumn{2}{c|}{$\rho\in[0.1,1]$}            & \multicolumn{2}{c}{$\rho=1$}                    & \multicolumn{2}{c}{$\rho\in[0.1,1]$}            \\
\midrule
Methods    & \multicolumn{1}{c}{MAE} & \multicolumn{1}{c}{RMSE} & \multicolumn{1}{c}{MAE} & \multicolumn{1}{c|}{RMSE} & \multicolumn{1}{c}{MAE} & \multicolumn{1}{c}{RMSE} & \multicolumn{1}{c}{MAE} & \multicolumn{1}{c}{RMSE} \\
\midrule
Cloud      & 1.176±0.007             & 1.426±0.007              & \multicolumn{1}{c}{/}   & \multicolumn{1}{c|}{/}    & 1.725±0.047             & 2.11±0.04                & \multicolumn{1}{c}{/}   & \multicolumn{1}{c}{/}    \\
Local      & 1.186±0.005             & 1.413±0.010              & \multicolumn{1}{c}{/}   & \multicolumn{1}{c|}{/}    & 1.735±0.015             & 2.119±0.020              & \multicolumn{1}{c}{/}   & \multicolumn{1}{c}{/}    \\
\midrule
FedAvg     & 1.415±0.006             & 1.635±0.007              & 1.411±0.011             & 1.628±0.013              & 1.862±0.008             & 2.247±0.004              & 1.867±0.006             & 2.252±0.006              \\
FedProx    & 1.427±0.006             & 1.645±0.008              & 1.413±0.017             & 1.630±0.017              & 1.873±0.011             & 2.260±0.011              & 1.865±0.013             & 2.257±0.011              \\
\midrule
Per-FedAvg & 1.217±0.007             & 1.434±0.009              & 1.239±0.026             & 1.456±0.029              & 1.733±0.010             & 2.114±0.014              & 1.763±0.023             & 2.148±0.021              \\
FedRep     & \textbf{1.147±0.009}    & \textbf{1.358±0.008}     & \textbf{1.143±0.011}    & \textbf{1.356±0.010}     & 1.648±0.008             & 2.022±0.006              & 1.623±0.018             & 1.990±0.021              \\
pFedMe     & 1.454±0.016             & 1.683±0.013              & 1.441±0.025             & 1.671±0.021              & 1.939±0.022             & 2.360±0.027              & 1.902±0.018             & 2.319±0.024              \\
Ditto      & 1.411±0.012             & 1.631±0.008              & 1.408±0.018             & 1.623±0.019              & 1.855±0.010             & 2.244±0.013              & 1.859±0.009             & 2.250±0.014              \\
APFL       & 1.419±0.005             & 1.637±0.006              & 1.415±0.008             & 1.636±0.009              & 1.876±0.008             & 2.264±0.009              & 1.883±0.016             & 2.271±0.016              \\
FedFomo    & 1.207±0.014             & 1.428±0.014              & 1.293±0.017             & 1.525±0.020              & 1.718±0.016             & 2.098±0.016              & 1.960±0.041             & 2.342±0.051              \\
FedALA     & 1.224±0.013             & 1.450±0.009              & 1.215±0.021             & 1.442±0.019              & 1.760±0.011             & 2.146±0.019              & 1.754±0.017             & 2.140±0.020              \\
\midrule
FedPAW     & 1.163±0.010             & 1.374±0.011              & 1.152±0.013             & 1.363±0.014              & \textbf{1.635±0.011}    & \textbf{2.003±0.014}       & \textbf{1.605±0.014}    & \textbf{1.978±0.016}               \\  
\bottomrule
\end{tabular}
\label{tab:performance}
\end{table*}

Table \ref{tab:performance} shows the prediction errors of FedPAW and all benchmark baselines at prediction horizons of 5 s and 10 s, demonstrating that FedPAW performs better than all baselines at 10 s, with a 0.8\% reduction in test MAE and a 0.9\% reduction in test RMSE (when $\rho=1$), and is only slightly overtaken by FedRep at 5 s. However, FedRep incurs a greater computational overhead than our FedPAW for clients . Due to non-IID data, the generalization ability of the global model is compromised, resulting in poor performance of both FedAvg and FedProx. Likewise, although Cloud has access to the most training data, the non-IID data makes it difficult for the model to converge sufficiently, and its performance is not the best due to the lack of personalization. Local performs worse than Cloud because it only trains on limited local data; although it adapts to the local data distribution of the client, it cannot utilize the knowledge learned by other clients. 

In PFL methods, Per-FedAvg reduces prediction errors through local fine-tuning, but fine-tuning focuses solely on local data and fails to acquire general information. FedRep outperforms, especially at the 5 s prediction horizon, because 
FedRep fine-tunes the head in each iteration while freezing the lower layers of the downloaded model, preserving most of the general information. However, the issue with personalized heads is that they also contain general information, and not sharing the head among clients leads to loss of the general information; moreover, fine-tuning the head requires additional time. Therefore, our FedPAW surpasses FedRep
at the 10 s prediction horizon because longer-term predictions require more general information. pFedMe and Ditto, which learn additional personalized models through a proximal term, perform poorly. For the vehicle speed prediction task, the effect of the proximal term appears to be poor, as evidenced also in FedProx. APFL performs poorly, as capturing the required information from the global model through a numerical mixing parameter is too trivial. FedFomo performs well, aggregating other relevant client models locally, and can limit useless models in federated updates to preserve the personalized performance of each model. FedALA performs better by adaptively learning aggregation weights on clients to capture the required information in the global model precisely. Our proposed FedPAW performs excellently by performing personalized aggregation on the server, precisely capturing client-specific information in the local models.

\begin{figure}[htpb]
    \centering
    \includegraphics[width=\linewidth]{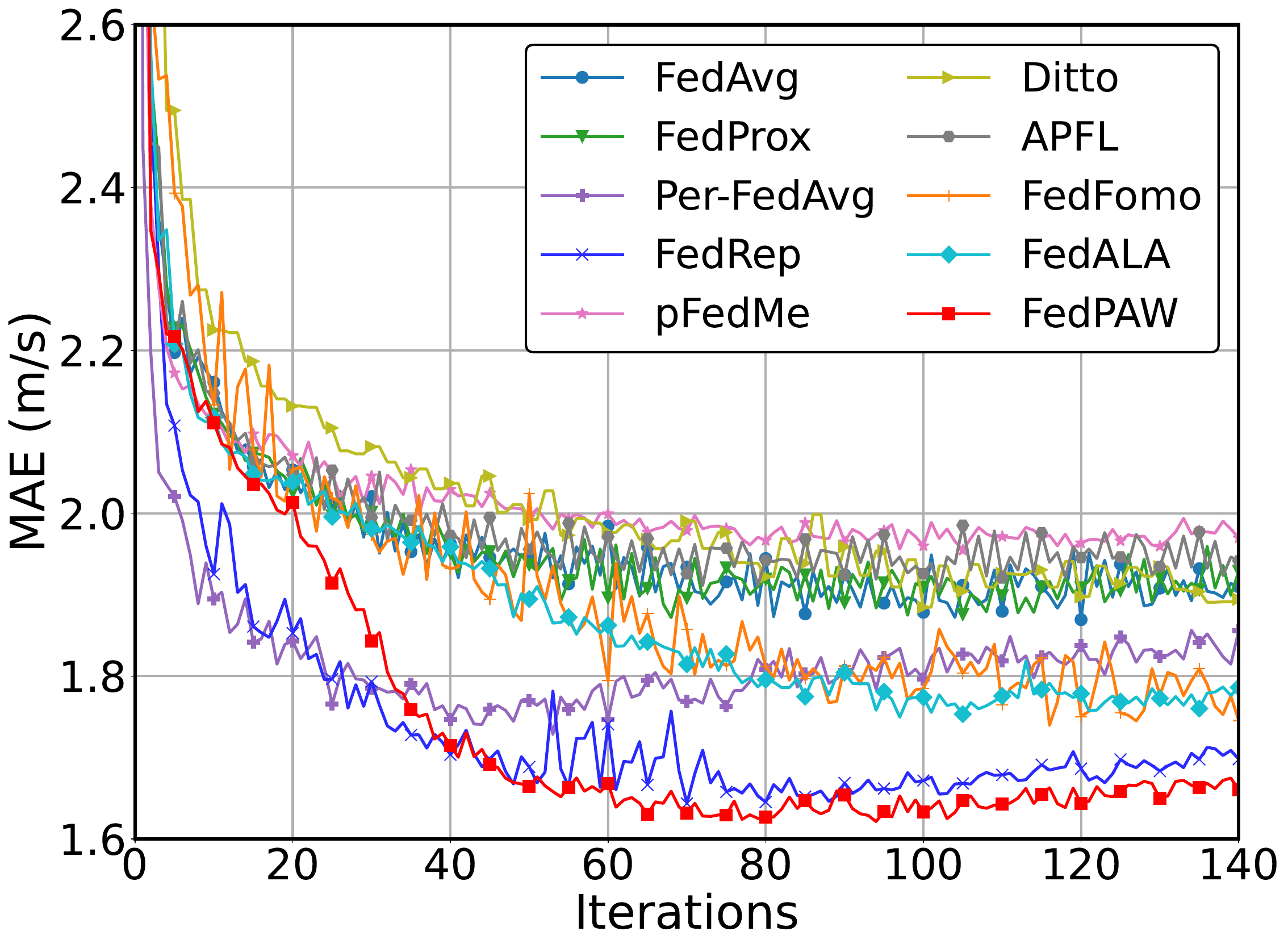}
    \caption{Test MAE (m/s) curves of different methods at a prediction horizon of 10 s.}
    \label{fig:mae_10s}
\end{figure}

Figure \ref{fig:mae_10s} presents the test MAE curves of FedPAW and all FL baseline at a prediction horizon of 10 s, demonstrating that FedPAW gradually converges and maintains the lowest prediction error after 60 iterations. In comparison, among the other methods, Per-FedAvg, FedFomo, and FedALA exhibit better performance, with FedRep performing the best besides of FedPAW. Per-FedAvg converges the fastest due to local fine-tuning, but this leads to significant overfitting in later iterations. FedFomo experiences considerable fluctuations in its MAE curve due to the instability of its aggregation strategy. FedRep, which fine-tunes the model head locally, converges quickly but lacks general information in the head, resulting in greater fluctuations in the early phase of the MAE curve. FedPAW  employs similar aggregation strategies as FedALA, starting with personalized aggregation, which converges slower initially compared with FedRep and Per-FedAvg. However, FedPAW benefits from a more stable convergence process and highest accuracy throughout, with FedPAW outperforming all other benchmarks after 60 iterations.

\subsection{Stability}
In real driving scenarios, partial of the vehicles may not continuously participate in the entire FL process due to reasons such as insufficient remained energy, lack of computing and storage resources, and unstable network conditions. We simulate this scenario by varying  joining ratio $\rho$ of the client in each iteration. Specifically, we uniformly sample a value of $\rho$ within a given range in each iteration. Table \ref{tab:performance} shows the prediction errors of FedPAW and all baselines under the extreme dynamic scenario of $\rho \in [0.1,1]$. Compared to $\rho=1$, it is evident that the stability of almost all methods decreases (as indicated by an increase in standard deviation). FedPAW still maintains its stability in this extreme dynamic scenario, with only a slight increase in standard deviation.

\begin{figure*}[htpb]
    \centering
    \includegraphics[width=\textwidth]{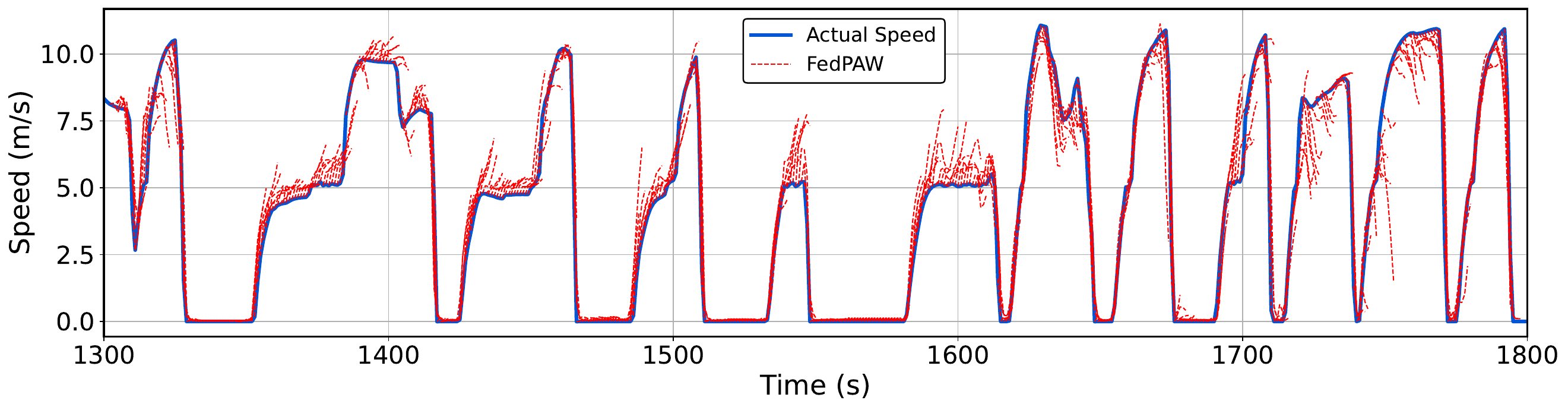}
    \caption{Speed prediction results of FedPAW framework on a subset of the test dataset at a prediction horizon of 5 s.}
    \label{fig:prediction_v}
\end{figure*}

\begin{table}[htpb]
\centering
\caption{The computation overhead for a prediction horizon of 10 s and communication overhead (parameters transmitted per iteration) when $\rho=1$. $\Sigma$ is the number of parameters in the backbone. $\alpha$ $(\alpha < 1)$ is the ratio of parameters of the feature extractor in the backbone in FedRep. $n$ $(n \geq 1)$ is the number of other clients each client receives in FedFomo.}

\begin{tabular}{l|cc|c}

\toprule
           & \multicolumn{2}{c|}{Computation} & Communication \\
\midrule
Methods    & Total time     & Time/iter.     & Param./iter.  \\
\midrule
FedAvg     & 623 s           & 4.66 s        & $2\ast\Sigma$           \\
FedProx    & 676 s           & 5.18 s          & $2\ast\Sigma$          \\
\midrule
Per-FedAvg & 782 s           & 10.12 s           & $2\ast\Sigma$             \\
FedRep     & 655 s           & 4.98 s          & $2\ast\alpha\ast\Sigma$             \\
pFedMe     & 4263 s          & 21.23 s       & $2\ast\Sigma$             \\
Ditto      & 819 s           & 8.69 s          & $2\ast\Sigma$              \\
APFL       & 1063 s          & 7.62 s        & $2\ast\Sigma$             \\
FedFomo    & 433 s           & 4.79 s          & $(1+n)\ast\Sigma$         \\
FedALA     & 797 s           & 5.23 s         & $2\ast\Sigma$            \\
\midrule
FedPAW     & 558 s           & 4.67 s          & $2\ast\Sigma$            \\
\bottomrule
\end{tabular}
\label{tab:overhead}
\end{table}

\subsection{Computation and Communication Overhead}
As shown in Table \ref{tab:overhead}, the experiment recorded the total computational time cost before convergence (determined by early stopping) for each FL method, comparing the computational overhead. The computation time per iteration for FedPAW is relatively low, only slightly higher than that of FedAvg. This is because FedPAW only adds the computational cost of the PA step on top of FedAvg, and the computation of aggregation weights is fast and does not require gradient descent. Additionally, PA is executed on the server, not imposing extra computational overhead on clients. Compared with FedPAW, benchmarks are beaten for numerous reasons. Per-FedAvg requires local model fine-tuning, and FedRep involves fine-tuning the model head, necessitating additional time. Learning personalized models in pFedMe involves extra training steps, leading to the highest computational cost per iteration, with Ditto facing a similar situation. APFL increases the computational overhead on clients due to freezing the global and local models to learn mixing parameters via gradient descent, and FedALA does the same for learning aggregation weights. FedFomo requires feeding data forward through downloaded client models to obtain approximate aggregate weights, which takes additional time. Additionally, the total computational time for FedPAW is also relatively low, indicating that fewer iterations are required for its convergence.

We theoretically compare the communication overhead for a client in a single iteration. As illustrated in Table \ref{tab:overhead}, most methods, similar to FedAvg, upload and download a single model in each iteration, leading to the same communication overhead. FedRep transmits only the lower layers of the model (feature extractor) in each iteration, resulting in the least communication overhead. FedFomo, downloading $n$ other client models in each iteration, incurs a high communication cost.

Therefore, FedPAW significantly reduces prediction error without increasing communication overhead, requiring only the server to undertake the computationally simple PA step additionally.

\subsection{Speed Prediction Results}

\begin{figure}[htpb]
\vspace{-0.05in}
    \centering
    \includegraphics[width=\linewidth]{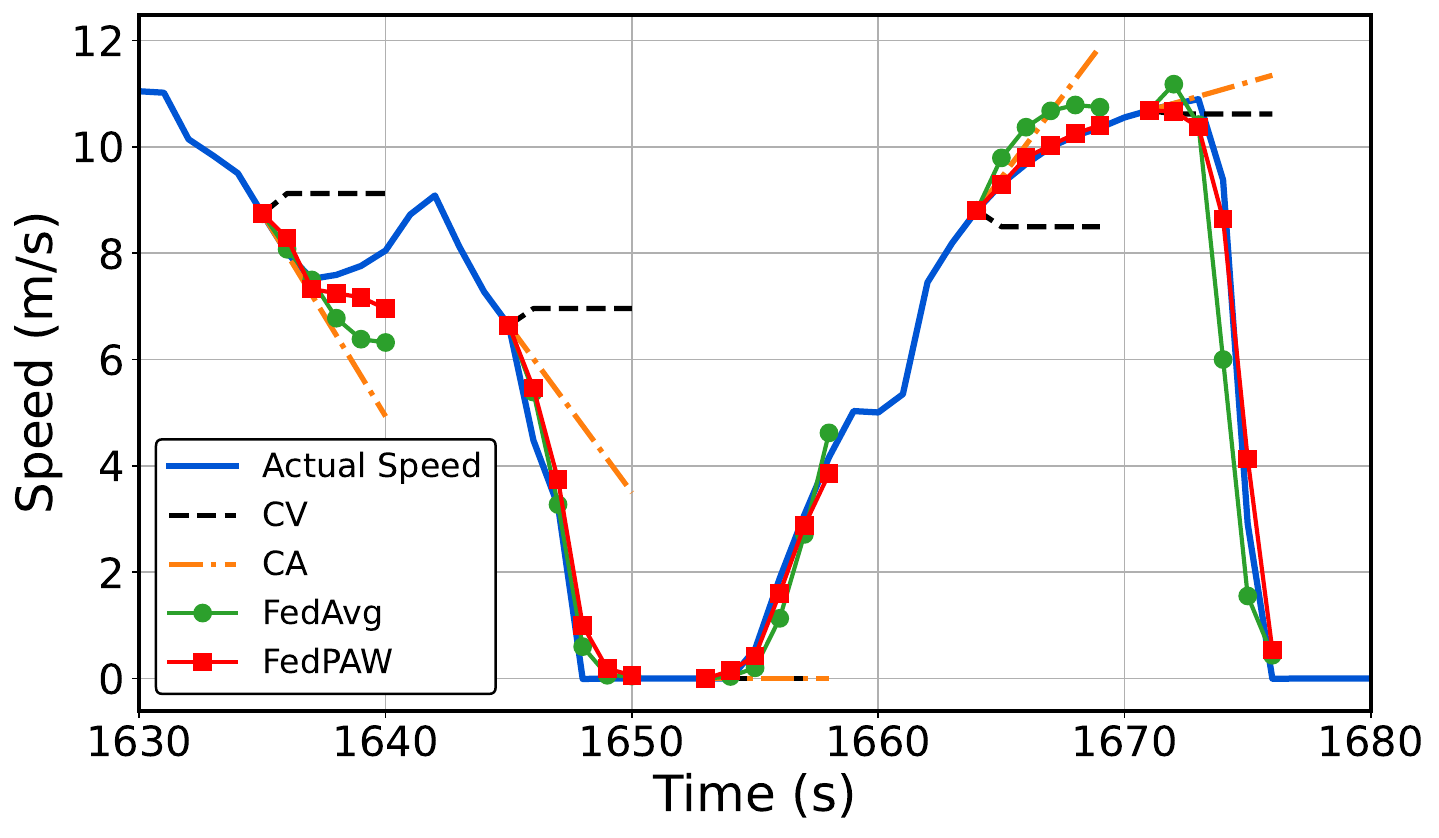}
    \caption{Speed prediction results results of the CV model, CA models, FedAvg, and FedPAW methods for five samples at a prediction horizon of 5 s.}
    \label{fig:prediction_v_part}
\end{figure}

Figure \ref{fig:prediction_v} shows the speed prediction results of the FedPAW framework on a subset of the test dataset for one client, evaluated every second at a prediction horizon of 5 s. It is evident that the predictive model within the FedPAW framework performs well in urban scenarios characterized by frequent starts and stops of vehicles, demonstrating strong generalization capabilities. To further analyze how the predictive model generalizes the movement of target vehicles interacting with preceding vehicles and traffic lights, Figure \ref{fig:prediction_v_part} presents a 50 s example sequence and compares the speed prediction results of five representative samples among actual speed, CV model, CA model,  FedAvg,  and FedPAW. The predictive effectiveness of the FedAvg and FedPAW  using the predictive model is much stronger than that of the CV and CA models, demonstrating the effectiveness of the proposed mechanism.

Specifically, in the first sample, with no traffic lights ahead and the target vehicle approaching a preceding vehicle, there is a slight deceleration; the predictive model infers this behavior through V2V communication, with FedPAW performing better. In the second sample, with a red traffic light ahead at a close distance, vehicles queuing, and the preceding vehicle changing lanes, the predictive model still performs well in this complex situation. In the third sample, with a traffic light changing from red to green in the next second, the predictive model accurately predicts the starting behavior based on V2I communication, with FedPAW performing slightly better. In the fourth sample, with no obstacles ahead but the target vehicle on a curve turning, the prediction model infers a slower acceleration based on the steering wheel angle feature, with FedPAW performing better. In the fifth sample, with a red traffic light ahead and no preceding vehicle, the target vehicle decelerates to a stop normally, and the predictive model predicts accurately, with FedPAW outperforming. Overall, the predictive performance of FedPAW is mostly superior to that of FedAvg, demonstrating the advantages of personalization in vehicle speed prediction.

\section{CONCLUSION}
\label{sec:conclusion}
In this paper, we propose FedPAW, a federated learning framework that performs personalized aggregation on the server to capture client-specific information for privacy-preserving personalized vehicle speed prediction, without imposing additional computational and communication overhead on clients. We introduce a new prediction model and have established a simulated driving dataset, CarlaVSP, through the CARLA simulator that distinguishes between different drivers and vehicle types. We demonstrate the efficacy of the FedPAW framework through extensive experiments on the CarlaVSP dataset, where FedPAW outperforms the eleven benchmark baselines in a comprehensive evaluation of prediction performance, stability, computational and communication overhead.


\bibliographystyle{IEEEtran}
\bibliography{IEEEabrv,references}


\vspace{11pt}

\begin{IEEEbiography}[{\includegraphics[width=1in,height=1.25in,clip,keepaspectratio]{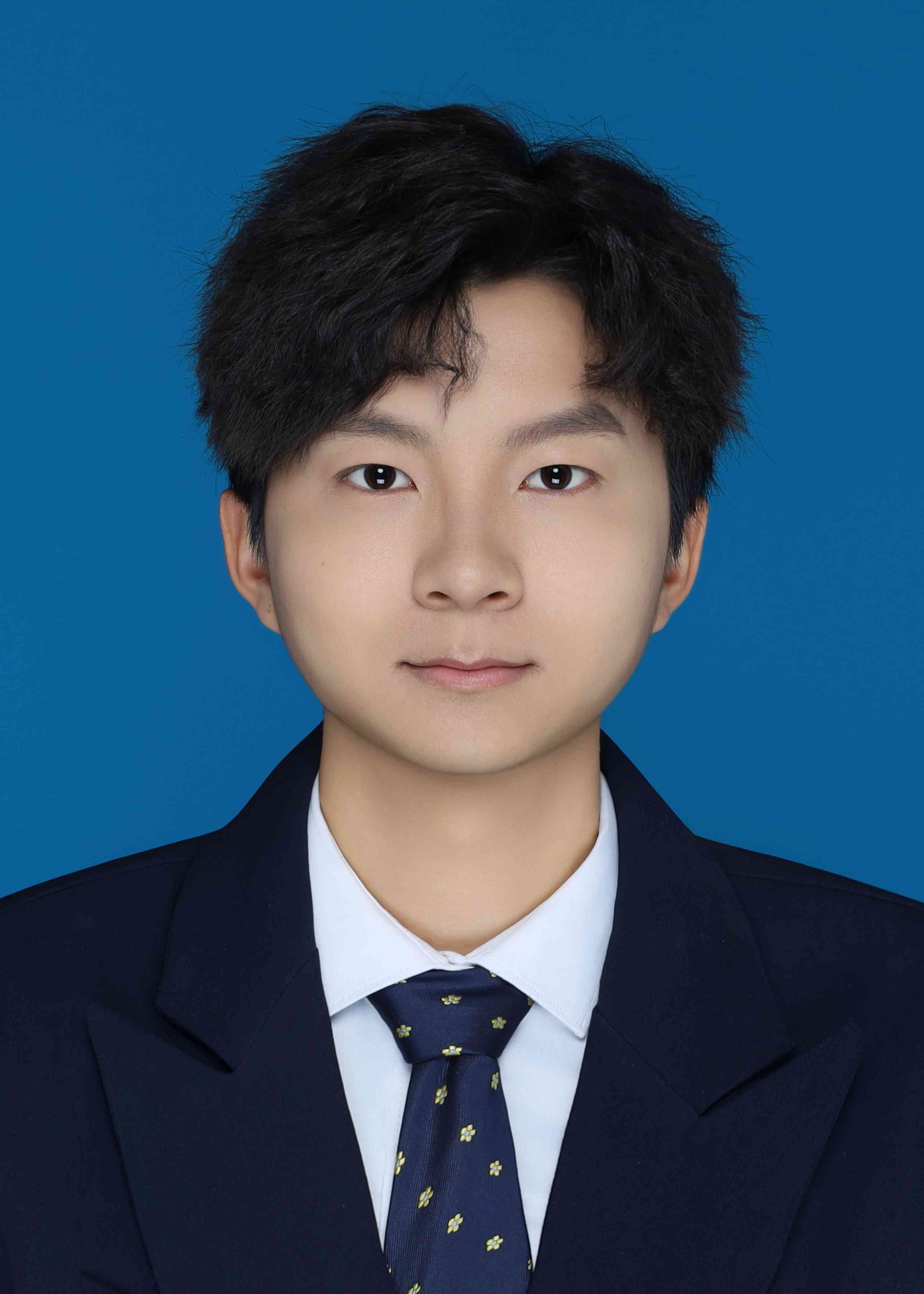}}]{Yuepeng He} received the bachelor's degree in software engineering from Nanchang University, in 2022. He is currently working toward the master's degree with Chongqing University. His research interests in personalized federated learning and internet of vehicles.
\end{IEEEbiography}

\begin{IEEEbiography}[{\includegraphics[width=1in,height=1.25in,clip,keepaspectratio]{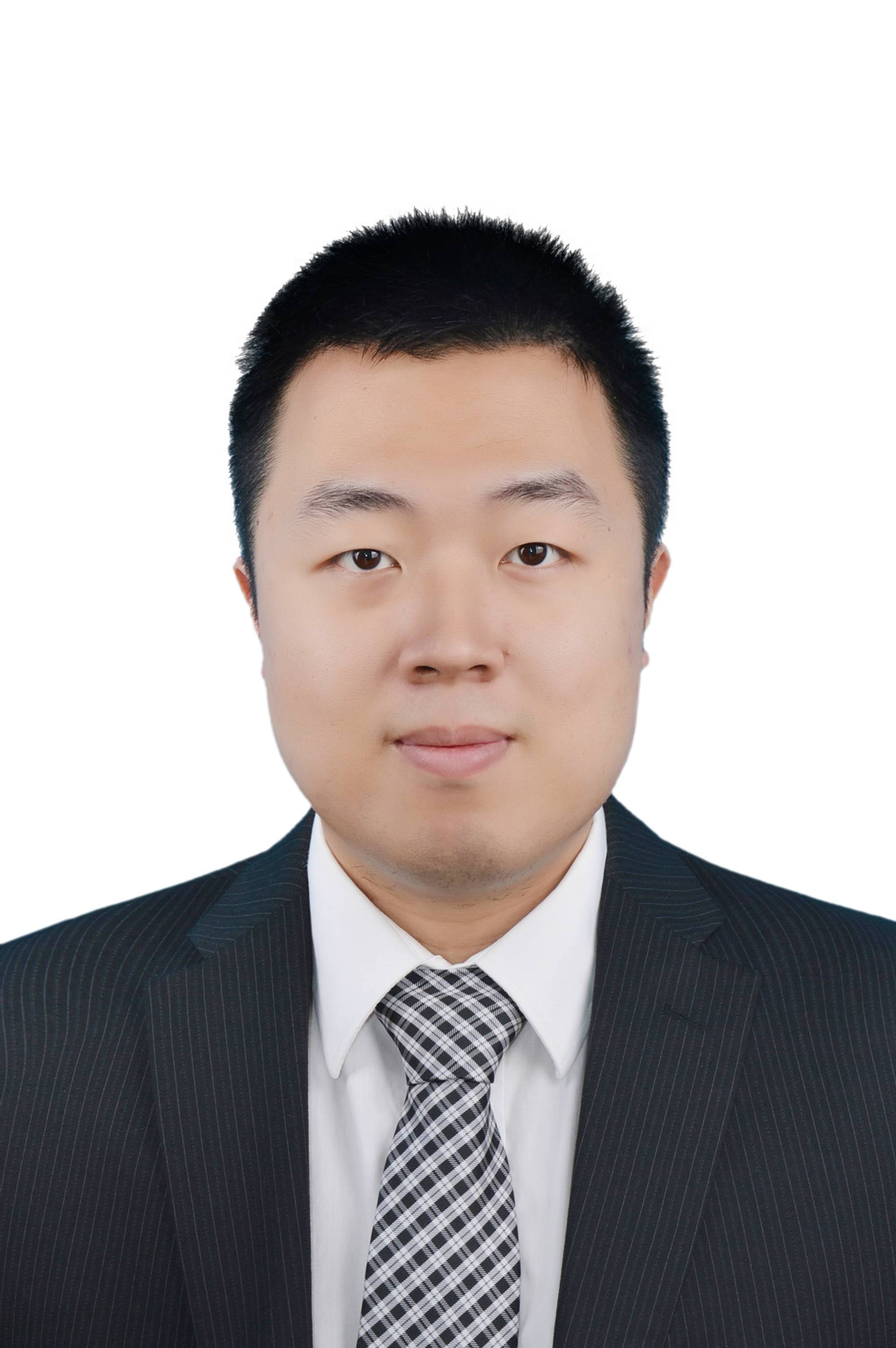}}]{Pengzhan Zhou} received the BS degree in both applied physics and applied mathematics from Shanghai Jiaotong University, Shanghai, China, and the PhD degree in electrical engineering from Stony Brook University, NY, USA. He is currently a professor with the College of Computer Science, Chongqing University, Chongqing, China. His research interests include Internet of Things, Internet of Vehicles, and Federated Learning.
\end{IEEEbiography}

\begin{IEEEbiography}[{\includegraphics[width=1in,height=1.25in,clip,keepaspectratio]{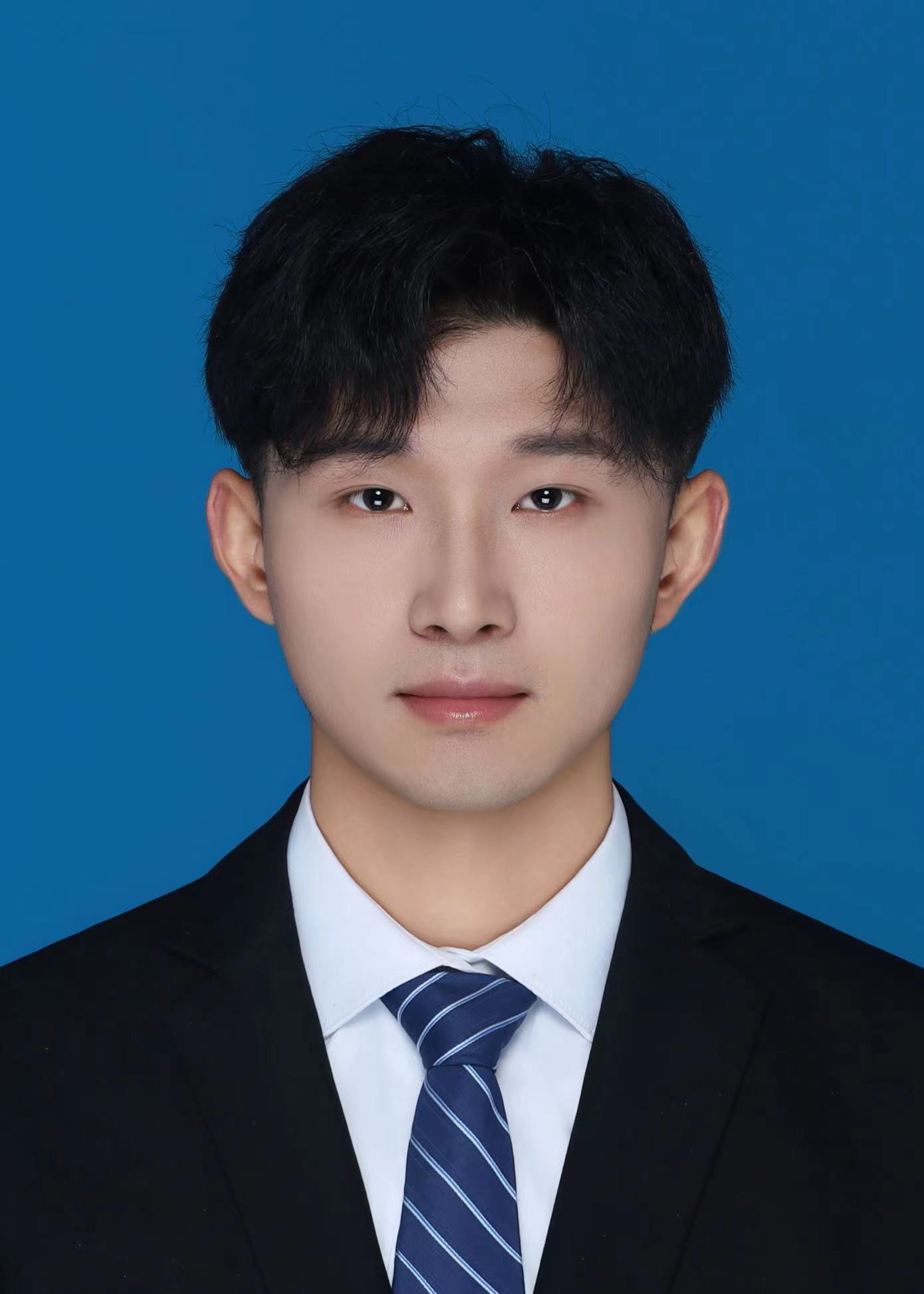}}]{Yijun Zhai} received the bachelor's degree in software engineering from Nanchang University, in 2022. He is currently working toward the master's degree with Chongqing University. His research interests in federated learning and internet of vehicles.
\end{IEEEbiography}

\begin{IEEEbiography}[{\includegraphics[width=1in,height=1.25in,clip,keepaspectratio]{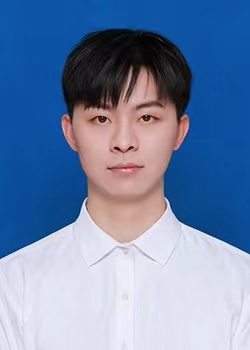}}]{Fang Qu} received the bachelor's degree in computer science from Chongqing University,in 2023. He is currently working toward the master's degree with Chongqing University. His research interests include Internet of Vehicle and computer vision.
\end{IEEEbiography}

\begin{IEEEbiography}[{\includegraphics[width=1in,height=1.25in,clip,keepaspectratio]{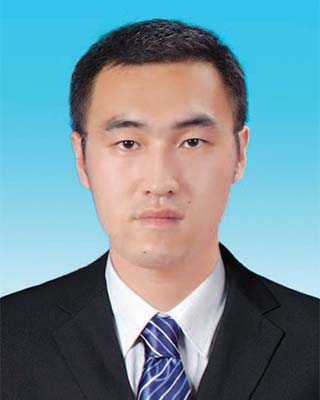}}]{Zhida Qin} received the B.S. degree in electronic information engineering from Huazhong University of Science and Technology, Wuhan, China, in 2014, and obtained Ph.D. degree in electronic engineering at Shanghai Jiao Tong University in 2020, Shanghai, China. He is currently an Assistant Professor with the School of Computer Science and Technology, Beijing Institute of Technology, Beijing, China. His current research interests include bandit learning, recommendation systems, and Internet of Things.
\end{IEEEbiography}

\begin{IEEEbiography}[{\includegraphics[width=1in,height=1.25in,clip,keepaspectratio]{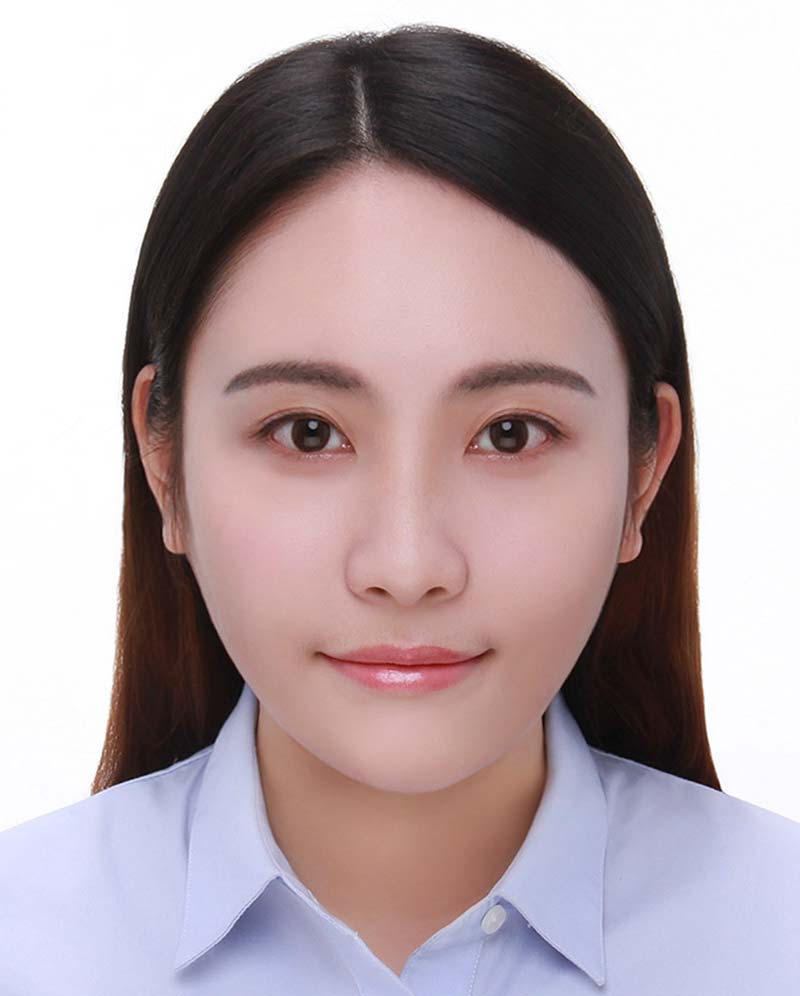}}]{Mingyan Li} received the B.E. degree in telecommunication engineering from Jilin University, Changchun, China, in 2015, and the Ph.D. degree from the Department of Electronic Engineering, School of Electronic Information and Electrical Engineering, Shanghai Jiao Tong University, Shanghai, China, in 2021. She was a Visiting Professor with the University of Waterloo, Canada, from 2019 to 2020. She joined the College of Computer Science, Chongqing University, in 2021, and currently works as a Post-Doctoral Research Associate. Her current research interests include industrial wireless networks and application in industrial automation, joint design of communication and control in industrial cyber-physical systems, software-defined networking and network slicing, ultra-reliable low-latency communication, and time-sensitive networks for industrial internet.
\end{IEEEbiography}

\begin{IEEEbiography}[{\includegraphics[width=1in,height=1.25in,clip,keepaspectratio]{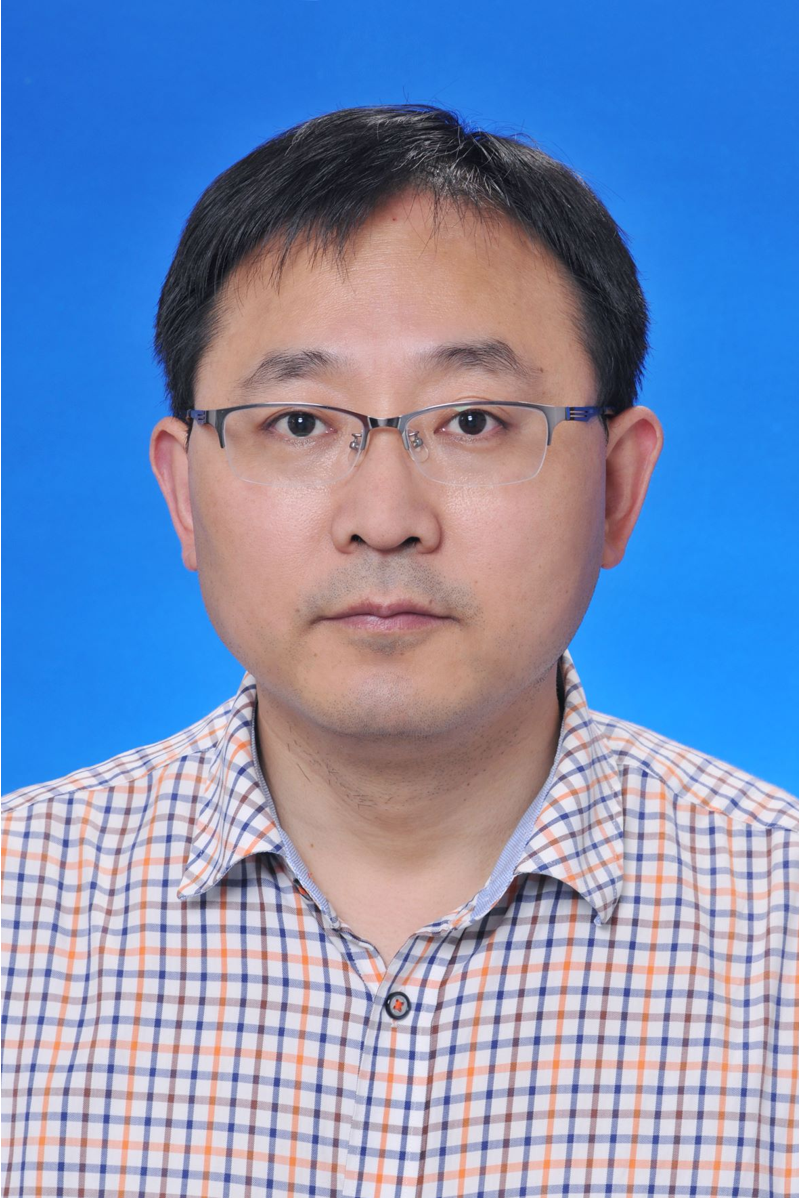}}]{Songtao Guo} (Senior Member, IEEE) received the BS, MS and PhD degrees in computer software and theory from Chongqing University, Chongqing, China, in 1999, 2003 and 2008, respectively. He was a professor from 2011 to 2012 with Chongqing University and a professor from 2013 to 2018 with Southwest University. Currently, he is a full professor with Chongqing University, China. He was a senior research associate with the City University of Hong Kong from 2010 to 2011, and a visiting scholar with Stony Brook University, New York, USA, from May 2011 to May 2012. His research interests include mobile edge computing, federated learning, wireless sensor networks, and wireless ad hoc networks. He has published more than 100 scientific papers in leading refereed journals and conferences. He has received many research grants as a principal investigator from the National Science Foundation of China and Chongqing and the Postdoctoral Science Foundation of China. He is a senior member of the ACM.
\end{IEEEbiography}

\vspace{11pt}

\vfill

\end{document}